\documentclass[10pt,twocolumn,letterpaper]{article}

\usepackage{iccv}
\usepackage{times}
\usepackage{epsfig}
\usepackage{graphicx}
\usepackage{amsmath}
\usepackage{amssymb}

\usepackage{url}            
\usepackage{booktabs}       
\usepackage{amsfonts}       
\usepackage{nicefrac}       
\usepackage{microtype}      

\usepackage{multirow}
\usepackage{bbm}
\usepackage{bm}
\usepackage{algorithm}
\usepackage{algorithmic}
\usepackage[table]{xcolor}
\usepackage{wrapfig}
\usepackage{setspace}
\usepackage{blindtext}
\usepackage{caption}
\usepackage{subcaption}
\usepackage{booktabs}

\usepackage[pagebackref=true,breaklinks=true,letterpaper=true,colorlinks,bookmarks=false]{hyperref}

\iccvfinalcopy 

\def\iccvPaperID{7184} 

\ificcvfinal\fi

\begin{document}

\title{Dense Interaction Learning for Video-based Person Re-identification}

\author{
	Tianyu He\textsuperscript{1}, Xin Jin\textsuperscript{2}, Xu Shen\textsuperscript{1}, Jianqiang Huang\textsuperscript{1}, Zhibo Chen\textsuperscript{2}, and Xian-Sheng Hua\textsuperscript{1}\\
	\textsuperscript{1}DAMO Academy, Alibaba Group \\
	\textsuperscript{2}University of Science and Technology of China\\
{\tt\small timhe.hty@alibaba-inc.com}
}

\maketitle
\ificcvfinal\thispagestyle{empty}\fi

\begin{abstract}
   Video-based person re-identification (re-ID) aims at matching the same person across video clips. Efficiently exploiting multi-scale fine-grained features while building the structural interaction among them is pivotal for its success. In this paper, we propose a hybrid framework, Dense Interaction Learning (DenseIL), that takes the principal advantages of both CNN-based and Attention-based architectures to tackle video-based person re-ID difficulties. DenseIL contains a CNN encoder and a Dense Interaction (DI) decoder. The CNN encoder is responsible for efficiently extracting discriminative spatial features while the DI decoder is designed to densely model spatial-temporal inherent interaction across frames. Different from previous works, we additionally let the DI decoder densely attends to intermediate fine-grained CNN features and that naturally yields multi-grained spatial-temporal representation for each video clip. Moreover, we introduce Spatio-TEmporal Positional Embedding (STEP-Emb) into the DI decoder to investigate the positional relation among the spatial-temporal inputs. Our experiments consistently and significantly outperform all the state-of-the-art methods on multiple standard video-based person re-ID datasets.
\end{abstract}

\section{Introduction}
\label{sec:intro}

Person re-identification (re-ID) tells whether a person-of-interest has been noticed in a different location by another camera. It is essential to many important surveillance applications such as tracking~\cite{wang2013intelligent} and retrieval~\cite{zheng2017sift}. In recent years, significant progresses have been achieved in image-based person re-ID~\cite{li2014deepreid,su2017pose,sun2018beyond,he2021partial}, as well as the video-based one~\cite{zheng2016mars,li2018diversity,yang2020spatial}, due to the rapid development of Convolutional Neural Networks (CNN)~\cite{krizhevsky2012imagenet}.

\begin{figure}[ht]
	\centering
	\begin{subfigure}{.45\textwidth}
		\centering
		\includegraphics[width=\textwidth]{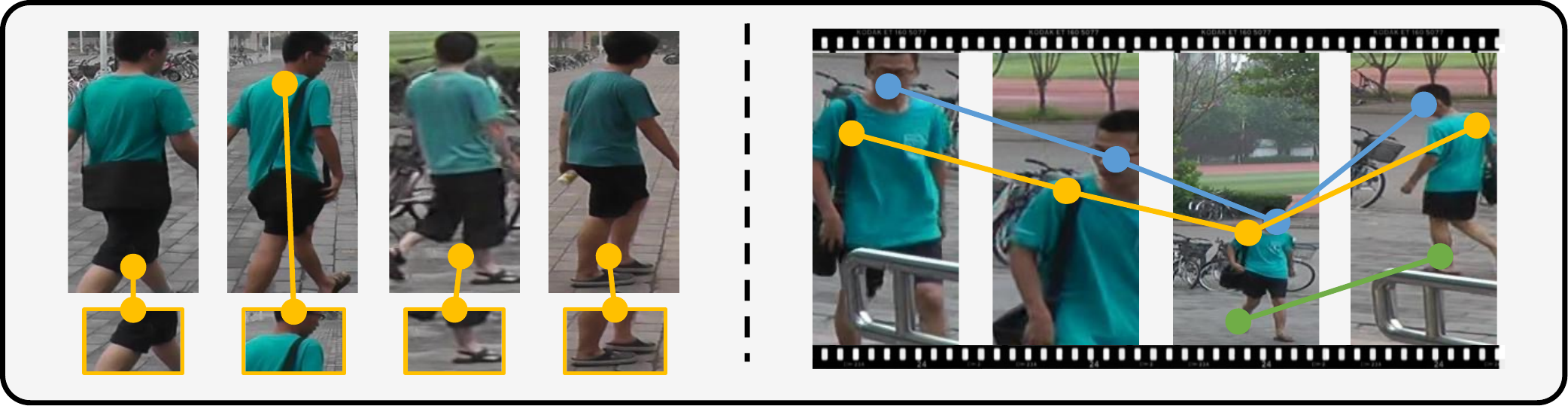}
		\caption{Left: different identities with similar appearance. Right: the same identity with misalignment or occlusion.}
		\label{fig:intro_challenge}
	\end{subfigure}%
	\vspace{2mm}
	\begin{subfigure}{.475\textwidth}
		\centering
		\includegraphics[width=\textwidth]{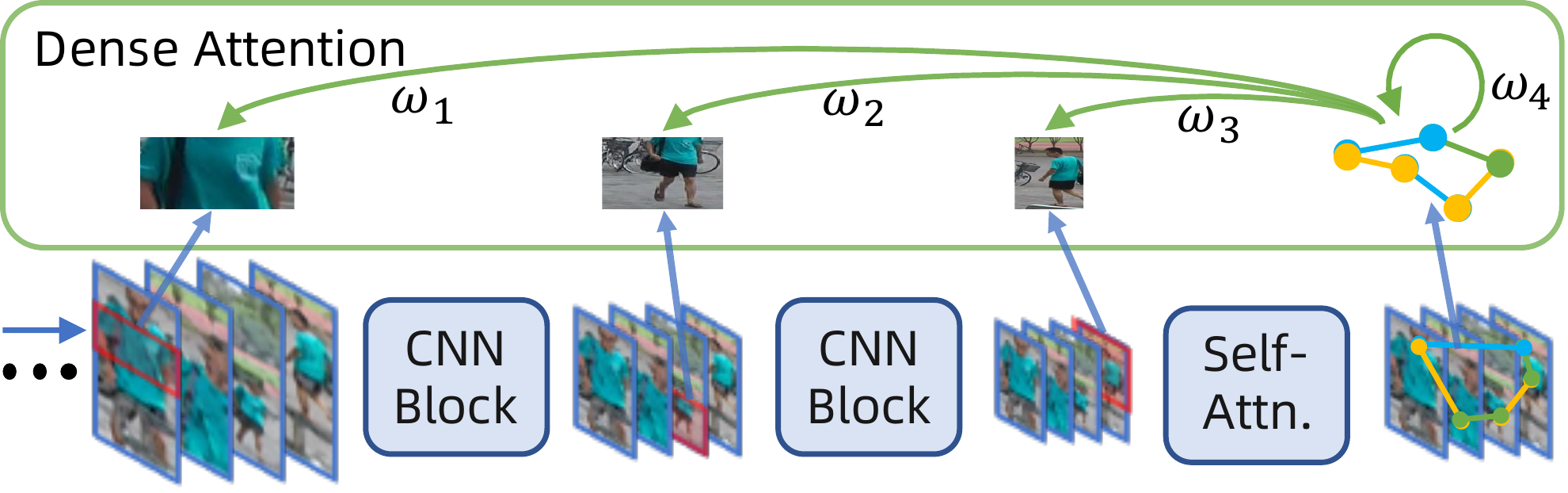}
		\caption{Our proposed Dense Interaction Learning (DenseIL).}
		\label{fig:intro_denseil}
	\end{subfigure}
	\vspace{-3mm}
	\caption{(a): Two key challenges existed in video-based person re-ID. (b): To tackle these two challenges, we introduce a DenseIL framework to densely capture multi-grained spatial-temporal interaction with the guidance of global relationship, where the preferences ($\omega$) are automatically learned by our proposed Dense Attention.}
	\vspace{-3mm}
\end{figure}

The goal of image-based person re-ID is to match person images captured in different times and locations. To achieve this goal, recent methods are proposed to better dig appearance features while concurrently to maintain the robustness with respect to body part misalignment~\cite{li2015multi,qian2017multi,chang2018multi,zhou2019omni,wei2017glad,sun2018beyond,wang2018learning,jin2020semantics,jin2020style,he2021partial}. Nevertheless, it is still insufficient to model the discrete relationships among various misaligned body parts from a single image. This motivates the exploration on video-based re-ID, which further takes adjacent frames of the captured person into consideration, and therefore the occluded or missed parts can be inferred. In this sense, the key point to video-based re-ID lies in designing an architecture that is suitable for temporal dynamics, such as leveraging optical flow~\cite{chen2020temporal}, RNN~\cite{zhou2017see} and 3D CNN~\cite{li2019multi}.

However, the aforementioned methods neglect the importance of spatial-temporal interaction between body parts within intra- and inter-frame, which limits their effectivity. As a result, the state-of-the-arts~\cite{yang2020spatial,yan2020learning} suggest that graph convolutional network~\cite{kipf2016semi} has the merit of modeling spatial-temporal dependencies and shows promising performance on video-based re-ID task. However, their models are built upon the \emph{coarse-grained} representation while leaving the \emph{fine-grained} information implied in each frame not fully exploited.
As demonstrated in Figure~\ref{fig:intro_challenge}, when different identities share similar appearance, depending on coarse-grained knowledge is not enough to distinguish the difference. Instead, fine-grained information (such as a shoe, a shoulder bag,~\etc.) plays an enormous role in re-identification.

Inspired by this observation, in this paper, we present Dense Interaction Learning (DenseIL) that not only builds \emph{spatial-temporal interaction} between body parts but also densely exploits \emph{fine-grained cues} (see Figure~\ref{fig:intro_denseil}).
Basically, DenseIL is composed of a CNN encoder and a Dense Interaction (DI) decoder. The CNN encoder exerts its advantage on efficiently encoding spatial context into discriminative features~\cite{lecun1998gradient}. Our CNN encoder consists of several CNN Blocks (\eg, Res-Block~\cite{he2016deep}, Dense-Block~\cite{huang2017densely}, SE-Block~\cite{hu2018squeeze},~\etc.) and therefore is capable of generating a set of hidden features from low-level/high-resolution to high-level/low-resolution.

Our DI decoder comprises stacked self-attention~\cite{vaswani2017attention}, feed-forward layer, layer normalization~\cite{ba2016layer} and a newly proposed Dense Attention.
Specifically, our decoder reuses the self-attention module explored in vanilla Transformer to deliberately model spatial-temporal inherent interaction across frames. After that, the subsequent Dense Attention module simultaneously attends to \emph{both} the outputs of the self-attention module and the intermediate fine-grained CNN features of the CNN encoder with the guidance of global relationship (\ie, the last self-attention outputs). Thus, the decoder can naturally generate a multi-grained spatial-temporal representation for each video clip.

In contrast to ResNet~\cite{he2016deep} and DenseNet~\cite{huang2017densely} that associate features of preceding layers through \emph{summation} or \emph{concatenation}, we create dense information flow between CNN and Attention mechanism with the proposed Dense Attention scheme. Intuitively, our Dense Attention is designed to better facilitate the coordination of CNN and Attention mechanism by extracting hybrid information from both the convolution and preceding self-attention features in a single attention function. Acting in this way, the model automatically and flexibly learns the preference on the fine-grained context (frame level, extracted by CNN) and global spatial-temporal interaction across frames (sequence level, modeled by self-attention) when re-identifying person-of-interest, which will be beneficial to scale-variations and misalignment problem, as a by-product of our model design.

Moreover, since the DI decoder is intrinsically permutation-invariant, we further propose Spatio-TEmporal Positional Embedding (STEP-Emb) to explicitly enhance the chronological relation in intra- and inter-frame. By incorporating STEP-Emb, our DI decoder is able to investigate the absolute or relative position of the spatial-temporal inputs.

Experiments show that our method significantly outperforms the state-of-the-arts on three video-based person re-ID datasets. In particular, we achieve the best performance of 87.0\% and 97.1\% mAP on large-scale MARS and DukeMTMC-VideoReID datasets respectively.

\section{Background}
\label{sec:background}

\subsection{Related Work}

\paragraph{Video-based Person Re-identification.}
Video-based person re-identification (re-ID)~\cite{zheng2016mars} targets on re-identifying a video clip of the person-of-interest from massive gallery candidates.
The majority of the video-based re-ID algorithms pay close attention to efficiently modeling temporal context and exploiting complementary information from various frames, by leveraging optical flow~\cite{mclaughlin2016recurrent,chung2017two,chen2020temporal}, RNN~\cite{yan2016person,zhou2017see,xu2017jointly,liu2019spatial}, pooling~\cite{zheng2016mars,wu2018exploit}, 3D CNN~\cite{qiu2017learning,li2019multi,gu2020appearance}, attention mechanism~\cite{li2018diversity,fu2019sta,liu2019spatially,li2019global,zhang2020multi} and graph convolutional network~\cite{yang2020spatial,yan2020learning}. For example, Mclaughlin \etal~\cite{mclaughlin2016recurrent} utilize optical flow, RNN and temporal pooling simultaneously to take long and short term temporal cues into account. Qiu \etal~\cite{qiu2017learning} and Li \etal~\cite{li2019multi} both employ 3D CNN to attain spatial-temporal representation for each video clip. However, both optical flow and 3D CNN are sensitive to spatial misalignment between adjacent video frames.

More recently, the rise of attention mechanism~\cite{bahdanau2015neural} convincingly demonstrates high capability of selectively focusing on specific parts of the input signal. Inspired by this, lots of studies learn the weight of spatial and temporal features separately from a static perspective~\cite{liu2017quality,zhou2017see,xu2017jointly,li2018diversity,fu2019sta,hou2019vrstc,li2019global,subramaniam2019co,hou2020temporal}. However, they do not take fully advantage of the relationships between spatial and temporal body parts across different frames, thus yielding limited capability of representation.

\vspace{-3.0mm}
\paragraph{Attention Mechanism in Computer Vision Community.}
The attention mechanism is first proven to be helpful in Neural Machine Translation~\cite{bahdanau2015neural,vaswani2017attention}, of which Transformer~\cite{vaswani2017attention} is the most famous one. Existing works in computer vision community leverage the success of Transformer mainly by replacing the convolutions with self-attention operation~\cite{parmar2019stand,wang2020axial}, or by deploying attention as an add-on to existing convolutional models in tasks like image classification~\cite{hu2018squeeze,hu2018gather,bello2019attention,hu2019local}, object detection~\cite{wang2018non,cao2019gcnet}, segmentation~\cite{fu2019dual,zhang2020feature}, generative networks~\cite{zhang2019self}, inpainting~\cite{zeng2020learning}, \etc.

The tale of attention mechanism still continues. People start to consider borrowing the entire Transformer architecture to jointly modeling vision-language representations~\cite{sun2019videobert,lu2019vilbert,su2019vl,rahman2020integrating,cornia2020meshed} or exploiting relations of the objects in image object detection~\cite{carion2020end,zhu2020deformable}.
While different from aforementioned studies that build the entire Transformer \emph{on the highest-level of} CNN spatial features, we only engage the decoder of Transformer, and replace the vanilla encoder-decoder attention with the proposed Dense Attention to pay attention to multi-grained CNN representations.

\vspace{-3.0mm}
\paragraph{Encoder-Decoder Framework with Skip Connections.}
The encoder-decoder framework is widely applied in the areas of language~\cite{cho2014learning,bahdanau2015neural}, speech~\cite{chan2016listen} and vision~\cite{long2015fully,zhu2017unpaired}. For image and video processing, Ronneberger \etal~\cite{ronneberger2015u} first introduce concatenation-style skip connections~\cite{he2016deep,huang2017densely} to the encoder-decoder-based fully convolutional networks~\cite{long2015fully}, and such framework is further utilized and refined in various computer vision tasks~\cite{mao2016image,fu2017dssd,tong2017image}.

In this work, instead of using the conventional summation-style or concatenation-style skip connections like above methods, we introduce Dense Attention to \emph{densely attend} to multi-grained features generated by the CNN encoder or the preceding decoder blocks.

\begin{figure*}
	\centering
	\begin{subfigure}{.3\textwidth}
		\centering
		\includegraphics[width=0.9\textwidth]{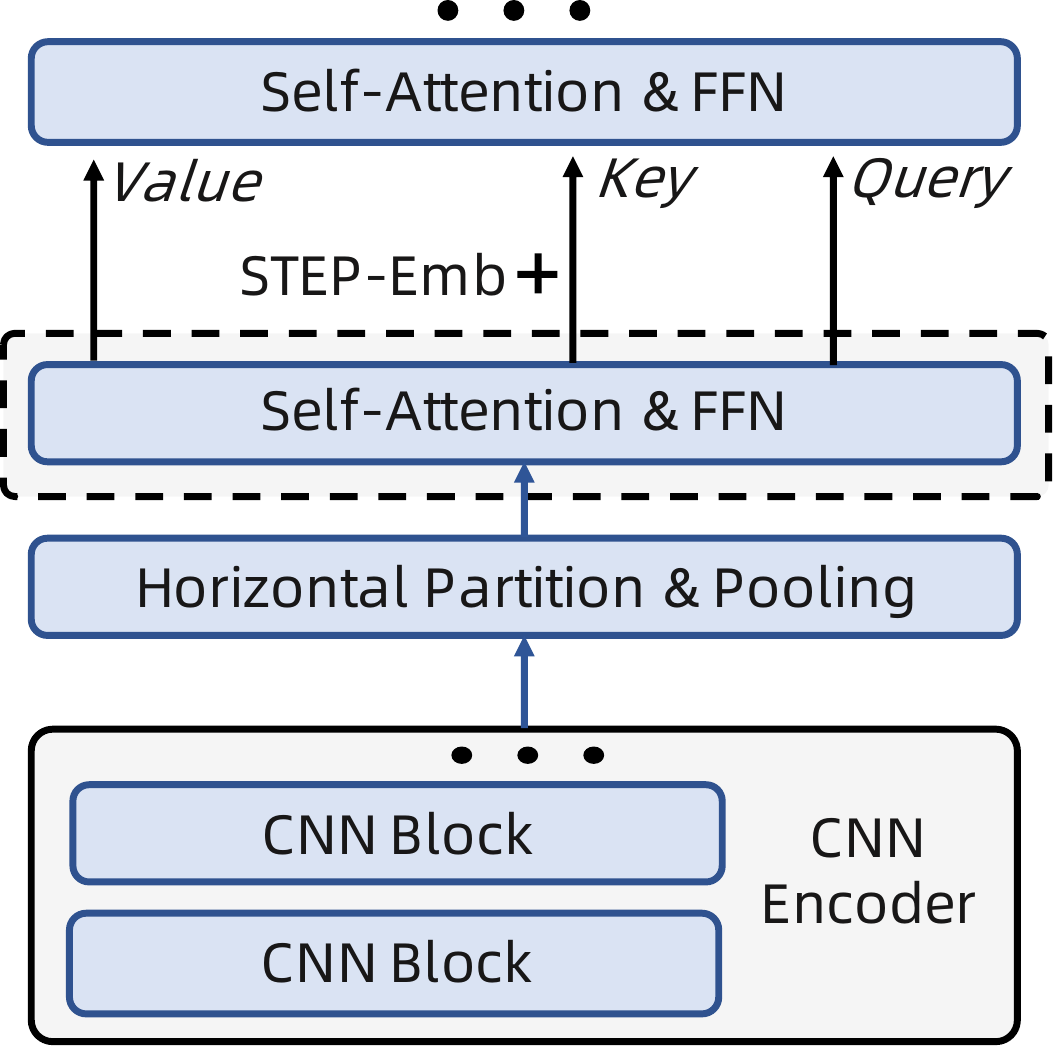}
		\caption{CNN-TransEnc}
		\label{fig:transenc}
	\end{subfigure}
	\begin{subfigure}{.3\textwidth}
		\centering
		\includegraphics[width=0.9\textwidth]{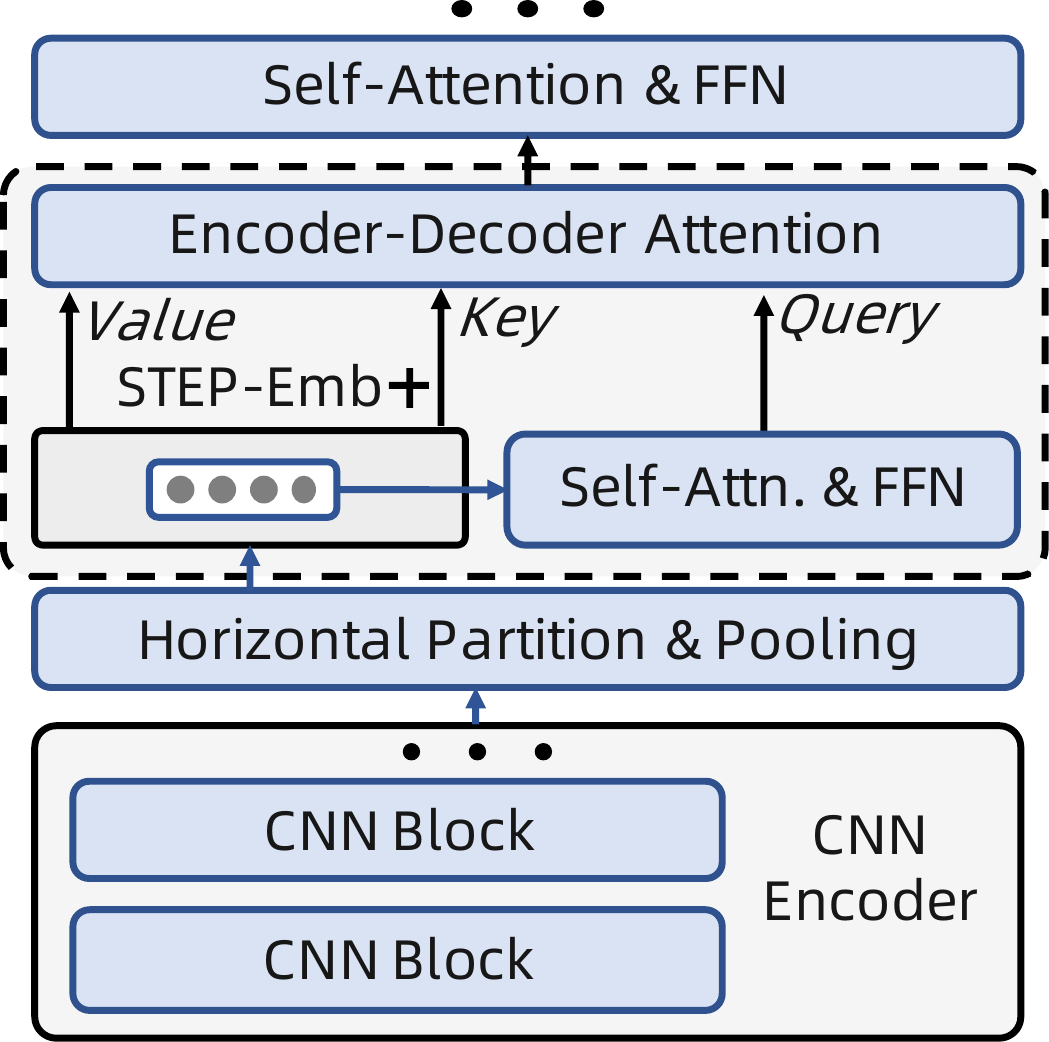}
		\caption{CNN-TransDec}
		\label{fig:transdec}
	\end{subfigure}
	\begin{subfigure}{.357\textwidth}
		\centering
		\includegraphics[width=0.9\textwidth]{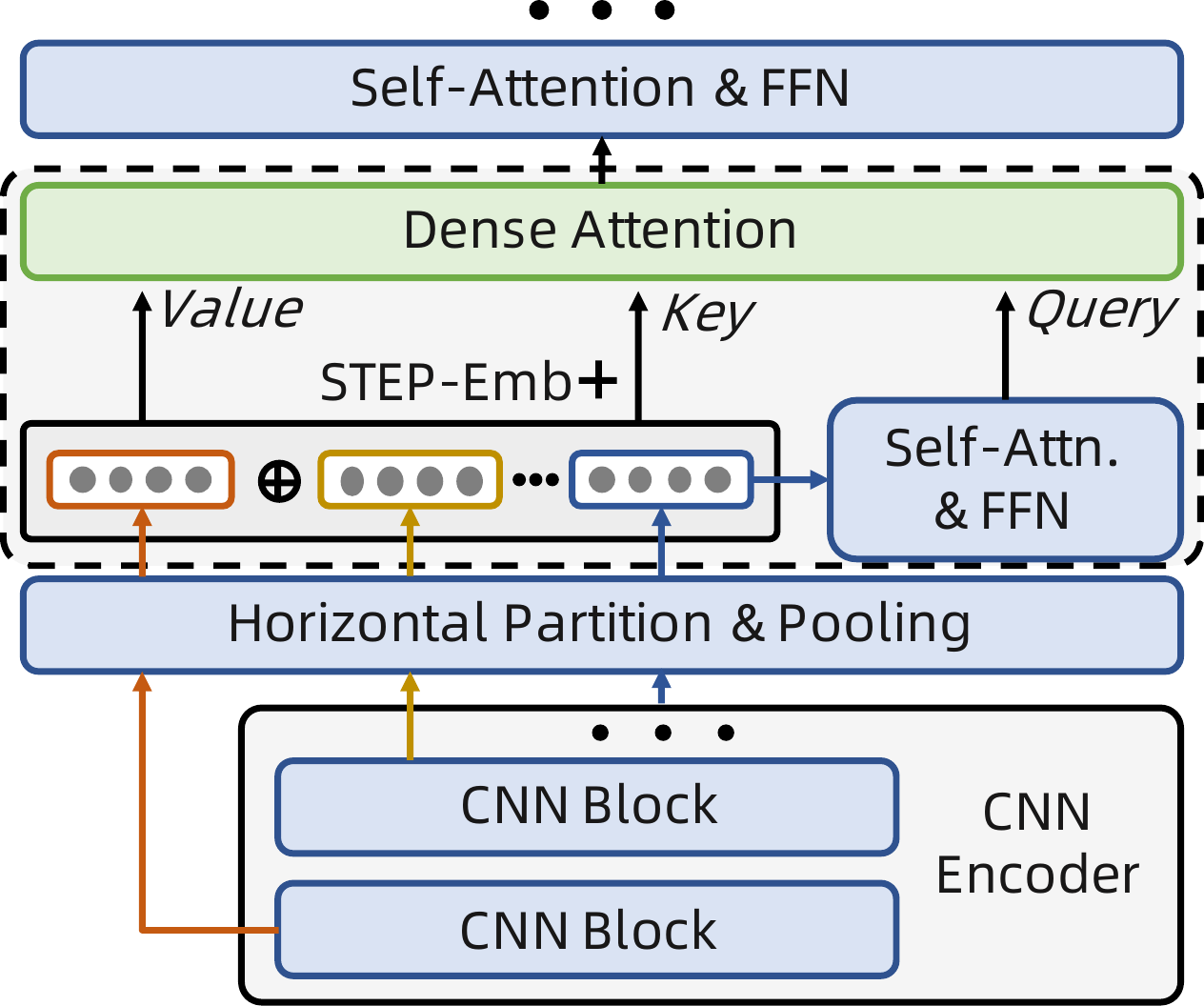}
		\caption{DenseIL}
		\label{fig:densetrans}
	\end{subfigure}
	\caption{The proposed three model variants for the video-based person re-ID task. (a) The decoder only consists of self-attention (is equivalent to the encoder of vanilla Transformer). (b) The decoder contains both self-attention and encoder-decoder attention (is equivalent to the decoder of vanilla Transformer). (c) Our DI decoder involves self-attention and the proposed Dense Attention (The $\oplus$ denotes the concatenation operation). All schemes are equipped with our proposed STEP-Emb. We omit the layer normalization for simplicity.}
	\vspace{-3mm}
	\label{fig:attndiff}
\end{figure*}

\subsection{Brief Introduction to Transformer}
\label{sec:transformer}

Transformer~\cite{vaswani2017attention} is a general encoder-decoder framework built upon self-attention and encoder-decoder attention that achieves state-of-the-art results on many language generation~\cite{he2018layer,wang2019multi,xia2019tied} and understanding tasks~\cite{devlin2019bert,yang2019xlnet}. It is a stacked architecture with several blocks. Each block composes two or three basic modules:

\vspace{-2mm}
\paragraph{A Self-Attention Module.}
Self-attention is commonly used to relate different positions of a single sequence and generate a weighted representation of its inputs. Formally, given an $I$-element set $\mathcal{X} = \{ x_1, x_2, \cdots, x_I \}$, the self-attention is defined as:
\begin{equation}  \label{eq:selfattn}
	\texttt{SelfAttn}(\boldsymbol{\textrm Q}, \boldsymbol{\textrm K}, \boldsymbol{\textrm V}) = \texttt{Softmax} (\frac{\boldsymbol{\textrm Q} \boldsymbol{\textrm K}^{\textbf{T}}}{\sqrt{d}}) \boldsymbol{\textrm V},
\end{equation}
where $\boldsymbol{\textrm Q}, \boldsymbol{\textrm K}, \boldsymbol{\textrm V}$ (\textit{query}, \textit{key}, \textit{value}) are all from $\mathcal{X}$ by a linear projection, and $d$ is the dimension of hidden states. Note that, the self-attention module can be employed in either an encoder or a decoder and is typically associated with a residual connection and layer normalization $\texttt{LN}$~\cite{ba2016layer}, resulting in the final output $\texttt{LN}(\mathcal{X} + \texttt{SelfAttn}(\boldsymbol{\textrm Q}, \boldsymbol{\textrm K}, \boldsymbol{\textrm V}))$.

\vspace{-2mm}
\paragraph{A Feed-Forward Layer.}
Feed-forward layer is analogous to activated fully-connected layer, that provides non-linearity to the model. For any $i \in \{1, \cdots, I\}$, $\texttt{FFN}(x_i) = w_2 \max (w_1 x_i + b_1, 0) + b_2$, where the $w$'s and $b$'s are the parameters to be learned. Similarly, this layer is followed with a residual connection and layer normalization.

\vspace{-2mm}
\paragraph{An Optional Encoder-Decoder Attention Module.}
To realize cross-lingual (\ie, source-to-target) learning, Vaswani \etal~\cite{vaswani2017attention} introduce an encoder-decoder attention that appears in the decoder only. Formally, let $\{ z_1, z_2, \cdots, z_I \}$ be the output of the last layer in the encoder, and $\boldsymbol{\textrm H}$ is the preceding hidden state in the decoder. The encoder-decoder attention is implemented similar to Equation~\eqref{eq:selfattn}, with the difference lies in that the \textit{query} ($\boldsymbol{\textrm Q}$) comes from $\boldsymbol{\textrm H}$, the \textit{key} ($\boldsymbol{\textrm K}$) and \textit{value} ($\boldsymbol{\textrm V}$) come from $\{ z_1, z_2, \cdots, z_I \}$. Eventually, this module outputs $\texttt{LN}(\boldsymbol{\textrm H} + \texttt{EncDecAttn}(\boldsymbol{\textrm Q}, \boldsymbol{\textrm K}, \boldsymbol{\textrm V}))$. More details can be found in the original paper~\cite{vaswani2017attention}.

\section{Dense Interaction Learning}
\label{sec:method}
In this section, we will give a detailed introduction to our proposed DenseIL. Our framework has a hybrid architecture and is composed of a CNN encoder and a Dense Interaction (DI) decoder. In general, with the assistance of the proposed Dense Attention, the DenseIL can enable the decoder densely attend to intermediate CNN features, naturally forming a multi-grained and interacted representation for each video clip. In the following, we first introduce each component we used in details and then give three variants for the overall architecture as demonstrated in Figure~\ref{fig:attndiff}.

\subsection{CNN Encoder}
In contrast to the vanilla Transformer that employs self-attention to construct hidden features for the source inputs, we use convolution-based transformation to generate hidden features for the inputs due to both its efficiency and high performance (\eg, translation equivariance and locality).

\vspace{-2mm}
\paragraph{Feature Extraction.}
Given a set of sampled video frames $\mathcal{X} = \{ x_1, x_2, \cdots, x_I \}$ with length $I$, the CNN encoder extracts the hidden spatial features block by block, where each block can be an arbitrary CNN structure (\eg, Res-Block~\cite{he2016deep}, Dense-Block~\cite{huang2017densely}, \etc.). We denote the spatial features in the $l$-th block as $\{ z_1^l, z_2^l, \cdots, z_I^l \}$.
In order to fully utilize the previously accumulated experience of pre-training~\cite{zheng2016mars}, we make minimal changes on the CNN architecture. Therefore, our CNN encoder can be initialized with ImageNet-pretrained parameters~\cite{deng2009imagenet}, endowing it with more power of robust representation.

\vspace{-2mm}
\paragraph{Horizontal Partition.}
Part-based re-ID model has enjoyed rich success in person re-ID~\cite{varior2016siamese,sun2018beyond,li2017learning,fu2019horizontal}, where each input sample is partitioned into patches by predefined priori knowledge or external supervision. The spirit of the part-based schemes lay on utilizing part-level features to provide more discriminative representation to distinguish the person-of-interest from others. Inspired by this, we horizontally divide the spatial features $\{ z_1, z_2, \cdots, z_I \}$ produced by CNN encoder into $P$ feature patches, and then perform average pooling on the divided feature to build a part-level feature vector $z_i^p \in \mathbb{R}^{1 \times d}$ for each patch, where $p \in \{1, \cdots, P\}$, and $d$ is the number of channels of the spatial features generated by CNN encoder. Eventually, we stack all feature vectors and obtain $\boldsymbol{\textrm Z} = [ z_1^1; \cdots; z_I^P ]$, where $\boldsymbol{\textrm Z} \in \mathbb{R}^{IP \times d}$ and $[ \cdot ]$ denotes the concatenation operation.

\subsection{Dense Interaction Decoder}  
As the CNN encoder generates the spatial features for each video clip, we further adopt the Dense Interaction (DI) decoder to model the long-range context, especially for excavating intra-frame and inter-frame relationships. In general, our DI decoder follows the design philosophy of the self-attention module and feed-forward layer in Transformer, but differs in the encoder-decoder attention module, which is substituted for the proposed Dense Attention module. The  main difference between the vanilla encoder-decoder attention and our Dense Attention is that ours allows the decoder to additionally attend to intermediate fine-grained CNN features, forming a hybrid dense connection between the multi-grained CNN and the attention features.

\vspace{-2mm}
\paragraph{Self-Attention.}
Our self-attention layer is developed with a focus on modeling the interaction between feature patches by reusing the innovations explored in prior works~\cite{parikh2016decomposable,vaswani2017attention,lin2017structured}. For the stacked feature vectors $\boldsymbol{\textrm Z}$, we conduct self-attention in Equation~\eqref{eq:selfattn}, where
\begin{equation}
	\boldsymbol{\textrm Q}, \boldsymbol{\textrm K}, \boldsymbol{\textrm V} = \boldsymbol{\textrm Z} \boldsymbol{\textrm W_q}, \boldsymbol{\textrm Z} \boldsymbol{\textrm W_k}, \boldsymbol{\textrm Z} \boldsymbol{\textrm W_v},
\end{equation}
$\boldsymbol{\textrm Z} \in \mathbb{R}^{IP \times d}$, and $\boldsymbol{\textrm W_q}, \boldsymbol{\textrm W_k}, \boldsymbol{\textrm W_v}$ are three learnable matrices to project $\boldsymbol{\textrm Z}$ into different spaces. In particular, we implement $\texttt{SelfAttn} (\cdot)$ with the multi-head attention as that proposed in Vaswani \etal~\cite{vaswani2017attention}. Our self-attention module is also associated with a residual connection and layer normalization $\texttt{LN} (\cdot)$ as described in Section~\ref{sec:transformer}.

Recent study~\cite{xiong2020layer} also shows that Pre-LN (put the $\texttt{LN} (\cdot)$ inside the residual connection) demonstrates stronger stability than the original Post-LN (place the $\texttt{LN} (\cdot)$ between the residual blocks). In this paper, we thereby adopt Pre-LN structure in our DI decoder.

\vspace{-2mm}
\paragraph{Feed-Forward Layer.}
We follow the spirit of the feed-forward layer $\texttt{FFN}( \cdot )$ of vanilla Transformer to endow the decoder with non-linearity.

\vspace{-2mm}
\paragraph{The Proposed Dense Attention.}
Residual networks~\cite{he2016deep,huang2017densely} build skip connections that can easily accumulate the features from the previous layer to the next layer, and can achieve great performance in a wide range of tasks. Inspired by this, we propose the Dense Attention, an operation that shares similar insights to the residual networks but realizes skip connections with attention mechanism, instead of summation~\cite{he2016deep} or concatenation~\cite{huang2017densely}.

Concretely, the Dense Attention simultaneously attends to the intermediate features $ \boldsymbol{\textrm Z}^l = \{ z_1^l, z_2^l, \cdots, z_I^l \}$ extracted by CNN, where $l \in \{ 1, \cdots, L \}$, and the high-level features $\boldsymbol{\textrm H}^r = \{ h_1^r, h_2^r, \cdots, h_{IP}^r \}$ generated from the preceding self-attention module (and the feed-forward layer) in the $r$-th block, where $r \in \{ 1, \cdots, R \}$. For the intermediate CNN features $ \boldsymbol{\textrm Z}^l $, which retain the fine-grained spatial information of inputs, we successively perform horizontal partition and an average pooling operation, denoted as $\texttt{PPool}(\cdot)$, to compress each partition into a feature vector before fed into Dense Attention. On the other hand, the features $\boldsymbol{\textrm H}^r$ are directly utilized, which represents the relationship between partitioned patches across frames. More precisely, our Dense Attention $\texttt{DenseAttn}(\boldsymbol{\textrm Q}, \boldsymbol{\textrm K}, \boldsymbol{\textrm V})$ is executed in a similar way as Equation~\eqref{eq:selfattn}, with modifications that the \textit{query} ($\boldsymbol{\textrm Q}$) comes from $\boldsymbol{\textrm H}^r$, the \textit{key} ($\boldsymbol{\textrm K}$) and \textit{value} ($\boldsymbol{\textrm V}$) comes from the concatenation of the pooled intermediate CNN features $ \texttt{PPool}(\boldsymbol{\textrm Z}^l) $ and the output of self-attention module $\boldsymbol{\textrm H}^r$:
\begin{equation}  \label{eq:denseattn2}
	\begin{split}
		\boldsymbol{\textrm V}^r & = [ \texttt{PPool}(\boldsymbol{\textrm Z}^1), \cdots, \texttt{PPool}(\boldsymbol{\textrm Z}^L), \boldsymbol{\textrm H}^r ] \boldsymbol{\textrm W_v}^r,  \\
		\boldsymbol{\textrm K}^r & = [ \texttt{PPool}(\boldsymbol{\textrm Z}^1), \cdots, \texttt{PPool}(\boldsymbol{\textrm Z}^L), \boldsymbol{\textrm H}^r ] \boldsymbol{\textrm W_k}^r,  \\
		\boldsymbol{\textrm Q}^r & = \boldsymbol{\textrm H}^r  \boldsymbol{\textrm W_q}^r,
	\end{split}
\end{equation}
where $[ \cdot ]$ represents the concatenation operation. The detailed distinction among self-attention, encoder-decoder attention and Dense Attention will be elaborated in Section~\ref{sec:overall}.

\subsection{Spatio-Temporal Positional Embedding}
\begin{figure}
	\centering
	\begin{subfigure}{.23\textwidth}
		\centering
		\includegraphics[width=0.3\textwidth]{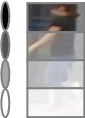}
		\caption{Spatial Pos. Emb.}
		\label{fig:spatialembed}
	\end{subfigure}
	\begin{subfigure}{.23\textwidth}
		\centering
		\includegraphics[width=0.479\textwidth]{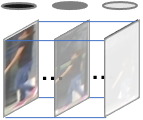}
		\caption{Temporal Pos. Emb.}
		\label{fig:temporalembed}
	\end{subfigure}
	\vspace{-1.5mm}
	\caption{The spatial and temporal positional embedding contribute to the proposed STEP-Emb together.}
	\vspace{-2.5mm}
	\label{fig:spembed}
\end{figure}

Since attention mechanism has no recurrent operation like RNN~\cite{hochreiter1997long} or convolution like CNN (\ie, permutation-variant), we need explicitly introduce the absolute or relative position of inputs to the model. Thus, we equip our DI decoder with positional embedding as demonstrated in Figure~\ref{fig:spembed}. The positional embedding is twofold: a spatial embedding and a temporal one. Both of them contribute to the final Spatial-Temporal Positional Embedding (STEP-Emb).

\vspace{-2mm}
\paragraph{Spatial Positional Embedding.}
As illustrated in Figure~\ref{fig:spatialembed}, the spatial positional embedding is constructed based on the partitioned spatial features, denoted as $z_i = \{ z_i^1, z_i^2, \cdots, z_i^P \}$ for $i$-th frame. We use sinusoidal embedding function to format embedding vectors~\cite{gu2018non,vaswani2017attention}: $\texttt{SpatialPos}(p, j)$ equals to $\sin(p / 10000^{j / d})$ if $j$ is even, and equals to $\cos(p / 10000^{j / d})$ otherwise. Here $p \in \{1, \cdots, P\}$ is the index of the partition, and $j \in \{1, \cdots, d\}$ is the index of the hidden dimension. For simplicity, we denote $s^p = \texttt{SpatialPos}(p, j)$.

\vspace{-2mm}
\paragraph{Temporal Positional Embedding.}
Compared with image-based recognition, video sequences additionally leak motion context for identifying pedestrians. However, Attention-based architectures (\ie, self-attention) are naturally permutation-invariant~\cite{vaswani2017attention}. Thus, they are incapable of explicitly modeling chronological relation between video frames, leading to a disruption of the intrinsic temporal structure. To tackle this problem, apart from the spatial positional embedding, we also build a temporal positional embedding to imply chronological order of inputs, as shown Figure~\ref{fig:temporalembed}. The temporal positional embedding can be formulated as : $\texttt{TemporalPos}(i, j)$ equals to $\sin(i / 10000^{j / d})$ if $j$ is even, and equals to $\cos(i / 10000^{j / d})$ otherwise. Here $i \in \{1, \cdots, I\}$ is the index of input frame. For simplicity, we denote $t_i = \texttt{TemporalPos}(i, j)$.

\vspace{-2mm}
\paragraph{Spatial-Temporal Positional Embedding.}
Based on the above two strategies, we derive the final Spatio-TEmporal Positional Embedding (STEP-Emb), which is constructed by the summation of the spatial and the temporal embedding. More precisely, the STEP-Emb for each feature patch $z_i^p$ can be represented as: $e_i^p = s^p + t_i$. Intuitively, the STEP-Emb improves the power of capturing long range interaction by enhancing the structural information between partitioned patches. We will demonstrate its ability in Section~\ref{exp:ablation}.

\subsection{Overall Architecture}
\label{sec:overall}
To dive deeply into the CNN-Attention hybrid structure, we introduce three model variants for the overall architecture. Figure~\ref{fig:attndiff} demonstrates the detailed configurations. For all the three variants, the components in the dashed boxes can be regarded as the basic building blocks to stack up. The final outputs of the decoder are processed by a batch normalization layer and a non-bias classifier layer.

\vspace{-2mm}
\paragraph{CNN encoder \& Vanilla Transformer Encoder (CNN-TransEnc).}
We directly cascade the CNN encoder, and the vanilla Transformer encoder, which mainly comprises self-attention module (\texttt{SelfAttn($\cdot$)}), to model interactions between feature partitions across frames. Therefore, the \textit{query}, \textit{key} and \textit{value} are all from the preceding self-attention module. The resulting architecture is illustrated in Figure~\ref{fig:transenc}.

\vspace{-2mm}
\paragraph{CNN encoder \& Vanilla Transformer Decoder (CNN-TransDec).}
We employ vanilla Transformer decoder combined with our CNN encoder as another model variant. As shown in Figure~\ref{fig:transdec}, compared with CNN-TransEnc, CNN-TransDec additionally includes the encoder-decoder attention (\texttt{EncDecAttn($\cdot$)}) to pay closer attention to the highest-level spatial features generated by the CNN encoder.

\vspace{-2mm}
\paragraph{CNN encoder \& Our DI Decoder with Dense Attention (DenseIL).}
The aforementioned CNN-TransEnc and CNN-TransDec only take the last-layer (the highest level) representations from the CNN encoder as inputs. As a comparison, our DenseIL (see Figure~\ref{fig:densetrans}) is equipped with the proposed Dense Attention, who is able to simultaneously attend to multi-scale intermediate spatial features in stacked CNN blocks and the corresponding interaction modeled by self-attention module.

\vspace{-2mm}
\paragraph{Loss Function.}
We adopt the simplest cross-entropy loss and batch triplet loss~\cite{hermans2017defense} for a fair comparison with the previous works~\cite{subramaniam2019co,hou2020temporal,gu2020appearance}. The two loss functions are calculated on the global average-pooled sequence-level feature vectors generated by the DI decoder.

\begin{table*}[t]
    \begin{subtable}[t]{0.32\textwidth}
		\footnotesize
		\caption{We compare Dense Attention with its counterparts.}
		\label{tab:ablationarch}
		{\def\arraystretch{1}\tabcolsep=0.85em
		\begin{tabular}[t]{@{}l|ccc@{}}
			\toprule[1.5pt]
			Methods & mAP & R-1 & R-5  \\
			\midrule
			Baseline                     & 82.1    & 87.3    & 95.6     \\
			\midrule
			CNN-TransEnc               & 85.7    &  89.4   &  96.6      \\
			CNN-TransDec           & 85.8    & 90.2    & 96.5      \\
			\midrule
			w/  Dense Concat.             & 86.4    & 89.8    & 96.8     \\
			w/  Dense Sum.             & 86.2    & 89.9    & 96.7   \\
			w/  Dense Attn.               & 87.0    & 90.8    & 97.1     \\
			\bottomrule[1.5pt]
		\end{tabular}
		}
	\end{subtable}\hspace{6pt}
	\begin{subtable}[t]{0.32\textwidth}
		\footnotesize
		\caption{DenseIL with / without positional embedding.}
		\label{tab:ablationemb}
		{\def\arraystretch{1}\tabcolsep=0.8em
		\begin{tabular}[t]{@{}l|ccc@{}}
			\toprule[1.5pt]
			Methods  & mAP & R-1 & R-5  \\
			\midrule
			Baseline                     & 82.1    & 87.3    & 95.6     \\
			\midrule
			DenseIL  & \multirow{2}{*}{86.3} & \multirow{2}{*}{89.9} & \multirow{2}{*}{97.0} \\
			w/o Pos. Emb. &  &  &   \\
			\midrule
			w/ Spatial-Emb.           & 86.6    & 90.5    & 97.1    \\
			w/ Temporal-Emb.             & 86.6    & 90.2    & 97.1    \\
			w/ STEP-Emb.               & 87.0    & 90.8    & 97.1    \\
			\bottomrule[1.5pt]
		\end{tabular}
		}
	\end{subtable}\hspace{6pt}
	\begin{subtable}[t]{0.32\textwidth}
		\footnotesize
		\caption{ Different number of blocks ($R$) with fixed $d = 2048$.}
		\label{tab:numofblocks}
		{\def\arraystretch{1.08}\tabcolsep=0.85em
		\begin{tabular}[t]{@{}l|cccc@{}}
			\toprule[1.5pt]
			\# Blocks   & \multicolumn{1}{c|}{GFs} & mAP & R-1 & R-5  \\
			\midrule
			Baseline                &  \multicolumn{1}{c|}{\multirow{2}{*}{16.39}}    & \multirow{2}{*}{82.1}    & \multirow{2}{*}{87.3}    & \multirow{2}{*}{95.6}     \\
			($R=0$) & \multicolumn{1}{c|}{}  &  &  &  \\
			\midrule
			$R=1$                  & \multicolumn{1}{c|}{17.43}     & 84.2    & 88.2    & 96.3    \\
			$R=2$                  & \multicolumn{1}{c|}{18.31}     & 86.3    & 89.6    & 96.9   \\
			$R=3$                  & \multicolumn{1}{c|}{19.18}     & 86.6    & 90.1    & 97.1    \\
			$R=4$                  & \multicolumn{1}{c|}{20.06}     & 87.0    & 90.8    & 97.1    \\
			\bottomrule[1.5pt]
		\end{tabular}
		}
	\end{subtable}
	\hfill
	
	\begin{subtable}[t]{0.32\textwidth}
		\vspace{6pt}
		\footnotesize
		\caption{Dense Attention with multi-grained CNN features. ($\boldsymbol{\textrm H}^r$ is omitted for simplicity).}
		\label{tab:howdense}
		{\def\arraystretch{1}\tabcolsep=0.65em
		\begin{tabular}[t]{@{}l|ccc@{}}
			\toprule[1.5pt]
			\textit{key} / \textit{value}  & mAP & R-1 & R-5  \\
			\midrule
			Baseline                     & 82.1    & 87.3    & 95.6     \\
			\midrule
			$\boldsymbol{\textrm Z}^4$ (CNN-TransDec)                    & 85.8    & 90.2    & 96.5     \\
			$\boldsymbol{\textrm Z}^4$ + $\boldsymbol{\textrm Z}^3$           & 86.7    & 90.6    & 97.1   \\
			$\boldsymbol{\textrm Z}^4$ + $\boldsymbol{\textrm Z}^3$ + $\boldsymbol{\textrm Z}^2$             & 87.0    & 90.8    & 97.1      \\
			$\boldsymbol{\textrm Z}^4$ + $\boldsymbol{\textrm Z}^3$ + $\boldsymbol{\textrm Z}^2$ + $\boldsymbol{\textrm Z}^1$               & 86.9    & 90.5    & 97.1   \\
			\bottomrule[1.5pt]
		\end{tabular}
		}
	\end{subtable}\hspace{6pt}
	\begin{subtable}[t]{0.32\textwidth}
		\vspace{6pt}
		\footnotesize
		\caption{Different number of dimensions of hidden states with fixed blocks $R=4$.}
		\label{tab:numofhidden}
		{\def\arraystretch{1.01}\tabcolsep=0.9em
		\begin{tabular}[t]{@{}l|cccc@{}}
			\toprule[1.5pt]
			\# Dims  & \multicolumn{1}{c|}{GFs} & mAP & R-1 & R-5 \\
			\midrule
			Baseline                  & \multicolumn{1}{c|}{16.39}   & 82.1    & 87.3    & 95.6      \\
			\midrule
			$d=256$                  & \multicolumn{1}{c|}{16.52}     & 86.6    & 90.1    & 97.1     \\
			$d=512$                  & \multicolumn{1}{c|}{16.75}     & 86.5    & 90.3    & 97.2   \\
			$d=1024$                  & \multicolumn{1}{c|}{17.48}     & 86.8    & 90.6    & 97.0    \\
			$d=2048$                  & \multicolumn{1}{c|}{20.06}     & 87.0    & 90.8    & 97.1     \\
			\bottomrule[1.5pt]
		\end{tabular}
		}
	\end{subtable}\hspace{6pt}
	\begin{subtable}[t]{0.32\textwidth}
		\vspace{6pt}
		\footnotesize
		\caption{We vary the number of partitions for each frame.}
		\label{tab:numofpart}
		{\def\arraystretch{1}\tabcolsep=1.3em
		\begin{tabular}[t]{@{}l|ccc@{}}
			\toprule[1.5pt]
			\# Partitions  & mAP & R-1 & R-5  \\
			\midrule
			Baseline                     & 82.1    & 87.3    & 95.6     \\
			\midrule
			$P=1$             & 86.1    & 89.7    & 96.9     \\
			$P=2$             & 86.8    & 90.5    & 97.1   \\
			$P=4$             & 87.0    & 90.8    & 97.1      \\
			$P=8$             & 86.6    & 90.5    & 97.0   \\
			\bottomrule[1.5pt]
		\end{tabular}
		}		
	\end{subtable}
	\hfill
	\vspace{-1mm}
	\caption{Ablation study on MARS dataset. GFs means GFLOPs. More details are explained in the text.}
	\vspace{-3mm}
    \label{tab:temps}
\end{table*}

\section{Experimental Results}
\label{sec:exps}

\subsection{Datasets and Evaluation Protocol}
\paragraph{Datasets.}
We evaluate our DenseIL on several commonly adopted video-based re-ID benchmarks, including MARS~\cite{zheng2016mars}, DukeMTMC-VideoReID (DukeV)~\cite{ristani2016performance,wu2018exploit} and iLIDS-VID~\cite{wang2014person}.

\paragraph{Evaluation Protocol.}
Following the common practices~\cite{zheng2015scalable,zheng2016mars,li2018diversity,yang2020spatial}, we resort to both Cumulative Matching Characteristic (CMC) curves at Rank-1 (R-1) to Rank-20 (R-20), and mean Average Precision (mAP) as evaluation metrics. We do not use re-ranking for all settings.

\subsection{Implementation Details}
\label{exp:expdetail}
We employ ImageNet-pretrained standard ResNet-50 as the initialization of our CNN encoder. To be comparable with the previous works~\cite{sun2018beyond,liu2019spatially,zhang2020multi,yang2020spatial}, we also remove the last spatial down-sampling operation in the $\texttt{conv5\_x}$ block for both the baseline and our schemes. Each input frame is resized to $256 \times 128$ with frame-level random horizontal flips~\cite{liu2019spatially} and sequence-level random erasing~\cite{zhong2020random,zhang2020multi,chen2020temporal} for data augmentation. We adopt a restricted random sampling strategy~\cite{li2018diversity,liu2019spatially,yang2020spatial} to randomly sample frames from equally divided $8$ chunks for each video clip. For each training batch, we randomly sample $16$ identities, each with $4$ tracklets~\cite{hermans2017defense,yang2020spatial}. We train our network for $800$ epoch with Adam optimizer for both the cross-entropy loss and the triplet loss~\cite{hermans2017defense}. The initial learning rate is set to $0.0002$ and is decayed by $10$ every $100$ epochs. The algorithm is implemented with PyTorch\footnote{Based on: \url{https://github.com/yuange250/not_so_strong_baseline_for_video_based_person_reID}.} and trained on a $4$-GPU machine.

\subsection{Study on Design Choices}
\label{exp:instantiations}


\begin{figure*}[ht]
	\centering
	\includegraphics[width=.9\linewidth]{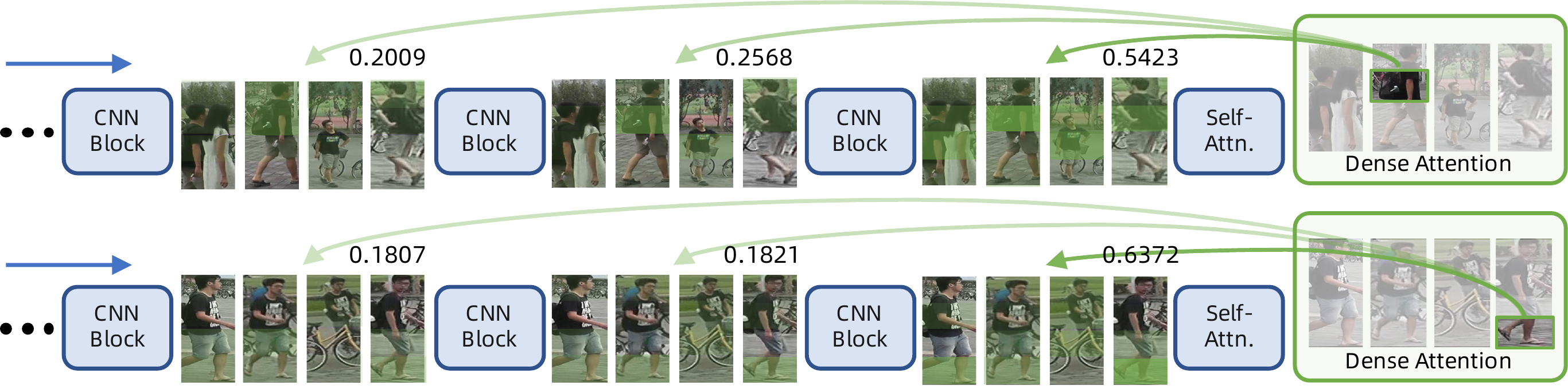}
	\caption{Visualization of the attention weights learned in Dense Attention. Darker color means higher attention weight.}
	\vspace{-3mm}
	\label{fig:vis}
\end{figure*}

\begin{figure}[ht]
	\centering
	\includegraphics[width=.95\linewidth]{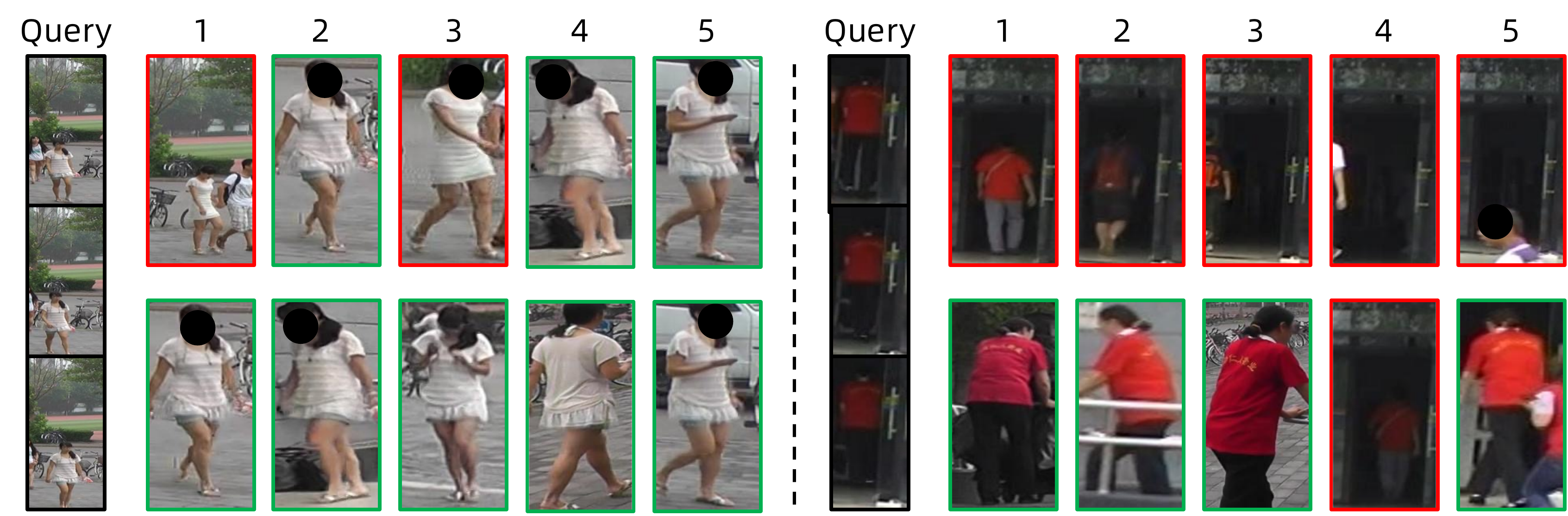}
	\caption{Visualization of re-identification results of the baseline (top) and DenseIL (bottom). The left column of each case is sampled frames of query sequence and the right five columns are top-5 retrieved sequences in the gallery set, where the item annotated with green box is correctly re-identified, and the red box denotes the wrong results.}
	\vspace{-3mm}
	\label{fig:vis_result}
\end{figure}

\paragraph{Spatial-Temporal Interaction Matters.}
We present the performance (\%) of CNN-TransEnc, CNN-TransDec and DenseIL (w/ Dense Attn.) on MARS dataset in Table~\ref{tab:ablationarch}, from which we can observe that solely allocating self-attention module to the decoder (CNN-TransEnc) already brings effective performance gain (+3.6\% mAP), while the one additionally inserted with encoder-decoder attention (CNN-TransDec) achieves comparable results (+3.7\% mAP). Such observation demonstrates the strong interaction-modeling capability of self-attention and exactly verifies one of our starting point: spatial-temporal interaction matters for video-based re-ID.

\vspace{-2.5mm}
\paragraph{Dense Attention \vs Dense Summation/Concatenation.}
Unlike ResNet~\cite{he2016deep} and DenseNet~\cite{huang2017densely} that combine features through summation or concatenation, we densely attend to preceding CNN and self-attention features by our proposed Dense Attention. Accordingly, we conduct experiments on densely summing (w/ Dense Sum.) or concatenating (w/ Dense Concat.) preceding CNN features to the decoder. To be more specific, we perform partition and pooling operation on the spatial features of each CNN block, and sum / concatenate them into the output of each self-attention module. We follow the same training strategy as described in Section~\ref{exp:expdetail} for both schemes and report mAP and CMC scores on MARS dataset in Table~\ref{tab:ablationarch}. As a result, the Dense Attention (w/ Dense Attn.) reaches the highest performance over the two counterparts (+0.6\% mAP). On the one hand, compared with summation or concatenation, our Dense Attention is able to adaptively learns the preference on low-level CNN features or high-level spatial-temporal interaction. On the other hand, the features learned by the CNN and the attention module may not lie in the same manifold, and concatenating them together can result in distribution gap. Instead, our Dense Attention enable better coordination between CNN and Attention by a relatively soft connection scheme.

\vspace{-2.5mm}
\paragraph{How Dense Attention Works?}
In order to investigate the functional principle of our Dense Attention, we adjust the composition of its \textit{key} and \textit{memory} in Equation~\eqref{eq:denseattn2} and conduct corresponding experiments. More precisely, let $\boldsymbol{\textrm Z}^1$, $\boldsymbol{\textrm Z}^2$, $\boldsymbol{\textrm Z}^3$ and $\boldsymbol{\textrm Z}^4$ be the intermediate features generated by four CNN blocks (ResNet-50), and $\boldsymbol{\textrm H}^r$ be the features generated from the preceding self-attention module. We allow Dense Attention to attend to $\boldsymbol{\textrm H}^r$ and part of $\{ \boldsymbol{\textrm Z}^1, \boldsymbol{\textrm Z}^2, \boldsymbol{\textrm Z}^3, \boldsymbol{\textrm Z}^4 \}$. The results on MARS dataset are demonstrated in Table~\ref{tab:howdense}. We can observe that when attending to $\boldsymbol{\textrm H}^r$ and any of $\{ \boldsymbol{\textrm Z}^2, \boldsymbol{\textrm Z}^3, \boldsymbol{\textrm Z}^4 \}$, DenseIL outperforms the CNN-TransDec counterpart by a large margin. It is also interesting to see that when further attending to the lowest-level CNN feature $\boldsymbol{\textrm Z}^1$, the performance slightly drops (-0.1\% mAP) compared with $\{ \boldsymbol{\textrm Z}^2, \boldsymbol{\textrm Z}^3, \boldsymbol{\textrm Z}^4 \}$. This might be due to the fact that the lowest-level CNN feature lacks discriminative information and is not conducive to the coordination with the attention scheme in terms of video re-ID task. For a better understanding of Dense Attention, we randomly feed image samples into DenseIL and map the attention weights of Dense Attention to colors. The results are visualized in Figure~\ref{fig:vis}, which suggests that the Dense Attention indeed cares about multi-grained CNN features simultaneously, while gives relatively more concern on the highest-level one. We also demonstrate the retrieval results in Figure~\ref{fig:vis_result}. We can observe that, in the left case, although there exists misalignment, movement and occlusion in the query respectively, our scheme is still able to match the person-of-interest accurately. In contrast, the baseline model misses the sequences of the same identity. In the right case, the baseline model re-identities the query incorrectly due to ignoring the fine-grained information between visually similar identities. In contrast, DenseIL captures the contour and the fine-grained characters on her back, yielding a satisfactory re-ID result.

\vspace{-2.5mm}
\paragraph{STEP-Emb Tackles Permutation-Invariant Issue.}
The philosophy behind STEP-Emb is to empower the DI decoder to capture long-range interaction by enhancing structural information between partitioned patches with embeddings. To verify its functionality, we remove the STEP-Emb (DenseIL w/o Pos. Emb.) and demonstrate its performance (\%) on MARS dataset in Table~\ref{tab:ablationemb}. It can be easily observed that employing either the Spatial Positional Embedding (w/ Spatial-Emb.) or the Temporal Positional Embedding (w/ Temporal-Emb) undeniably boosts the person re-ID performance (+0.3\% mAP). As the joint one, STEP-Emb, achieves the best performance consistently on all metrics. This suggests that STEP-Emb is indispensable for spatial-temporal positional modeling in our DI decoder.

\begin{table*}[t]
	\centering
	\renewcommand*{\arraystretch}{1.05}
	\small
	\begin{tabular}{@{}lcccccccccccc@{}}
	\toprule[1.5pt]
	\multicolumn{1}{c}{\multirow{2}{*}{Methods}} & \multirow{2}{*}{Proc.} & \multirow{2}{*}{Backbone} & \multicolumn{4}{c}{MARS} & \multicolumn{4}{c}{DukeV} & \multicolumn{2}{c}{iLIDS-VID}  \\ \cmidrule(l){4-13} 
	\multicolumn{1}{c}{} &  &  & mAP & R-1 & R-5 & \multicolumn{1}{c|}{R-20} & mAP & R-1 & R-5 & \multicolumn{1}{c|}{R-10} & R-1 & \multicolumn{1}{c}{R-5}  \\ \midrule
	\multicolumn{1}{l|}{STAN~\cite{li2018diversity}} & \multicolumn{1}{c|}{CVPR18} & \multicolumn{1}{c|}{Res50} & 65.8 & 82.3 & - & \multicolumn{1}{c|}{-} & - & - & - & \multicolumn{1}{c|}{-} & 80.2 & \multicolumn{1}{c}{-}  \\
	\multicolumn{1}{l|}{Snippet~\cite{chen2018video}} & \multicolumn{1}{c|}{CVPR18} & \multicolumn{1}{c|}{Res50} & 76.1 & 86.3 & 94.7 & \multicolumn{1}{c|}{98.2} & - & - & - & \multicolumn{1}{c|}{-} & 85.4 & \multicolumn{1}{c}{96.7} \\
	\multicolumn{1}{l|}{STA~\cite{fu2019sta}} & \multicolumn{1}{c|}{AAAI19} & \multicolumn{1}{c|}{Res50} & 80.8 & 86.3 & 95.7 & \multicolumn{1}{c|}{-} & 94.9 & 96.2  & 99.3  & \multicolumn{1}{c|}{99.6} & - & \multicolumn{1}{c}{-}  \\
	\multicolumn{1}{l|}{ADFD~\cite{zhao2019attribute}} & \multicolumn{1}{c|}{CVPR19} & \multicolumn{1}{c|}{Res50} & 78.2 & 87.0 & 95.4 & \multicolumn{1}{c|}{98.7} & - & - & - & \multicolumn{1}{c|}{-} & 86.3 & \multicolumn{1}{c}{97.4}  \\
	\multicolumn{1}{l|}{VRSTC~\cite{hou2019vrstc}} & \multicolumn{1}{c|}{CVPR19} & \multicolumn{1}{c|}{Res50} & 82.3 & 88.5 & 96.5 & \multicolumn{1}{c|}{-} & 93.5 & 95.0  & 99.1  & \multicolumn{1}{c|}{99.4} & 83.4 & \multicolumn{1}{c}{95.5}  \\
	\multicolumn{1}{l|}{GLTR~\cite{li2019global}} & \multicolumn{1}{c|}{ICCV19} & \multicolumn{1}{c|}{Res50} & 78.5 & 87.0 & 95.8 & \multicolumn{1}{c|}{98.2} & 93.7 & 96.3 & 99.3  & \multicolumn{1}{c|}{-} & 86.0 & \multicolumn{1}{c}{\textbf{98.0}}  \\
	\multicolumn{1}{l|}{COSAM~\cite{subramaniam2019co}} & \multicolumn{1}{c|}{ICCV19} & \multicolumn{1}{c|}{SE-Res50} & 79.9 & 84.9 & 95.5 & \multicolumn{1}{c|}{97.9} & 94.1 & 95.4  &  99.3 & \multicolumn{1}{c|}{-} & 79.6 & \multicolumn{1}{c}{95.3}  \\
	\multicolumn{1}{l|}{STE-NVAN~\cite{liu2019spatially}} & \multicolumn{1}{c|}{BMVC19} & \multicolumn{1}{c|}{Res50-NL} & 81.2 & 88.9 & - & \multicolumn{1}{c|}{-} & 93.5 & 95.2 & - & \multicolumn{1}{c|}{-} & - & \multicolumn{1}{c}{-}  \\
	\multicolumn{1}{l|}{MG-RAFA~\cite{zhang2020multi}} & \multicolumn{1}{c|}{CVPR20} & \multicolumn{1}{c|}{Res50} & 85.9 & 88.8 & 97.0 & \multicolumn{1}{c|}{98.5} & - & - & - & \multicolumn{1}{c|}{-} & 88.6 & \multicolumn{1}{c}{\textbf{98.0}}  \\
	\multicolumn{1}{l|}{MGH~\cite{yan2020learning}} & \multicolumn{1}{c|}{CVPR20} & \multicolumn{1}{c|}{Res50-NL} & 85.8 & 90.0 & 96.7 & \multicolumn{1}{c|}{98.5} & - & - & - & \multicolumn{1}{c|}{-} & 85.6 & \multicolumn{1}{c}{97.1}  \\
	\multicolumn{1}{l|}{STGCN~\cite{yang2020spatial}} & \multicolumn{1}{c|}{CVPR20} & \multicolumn{1}{c|}{Res50} & 83.7 & 90.0 & 96.4 & \multicolumn{1}{c|}{98.3} & 95.7 & 97.3 & 99.3 & \multicolumn{1}{c|}{-} & - & \multicolumn{1}{c}{-}  \\
	\multicolumn{1}{l|}{TCLNet~\cite{hou2020temporal}} & \multicolumn{1}{c|}{ECCV20} & \multicolumn{1}{c|}{Res50-TCL} & 85.1 & 89.8 & - & \multicolumn{1}{c|}{-} & 96.2 & 96.9 & - & \multicolumn{1}{c|}{-} & 86.6 & \multicolumn{1}{c}{-}  \\
	\multicolumn{1}{l|}{AP3D~\cite{gu2020appearance}} & \multicolumn{1}{c|}{ECCV20} & \multicolumn{1}{c|}{AP3D} & 85.1 & 90.1 & - & \multicolumn{1}{c|}{-} & 95.6 & 96.3 & - & \multicolumn{1}{c|}{-} & 86.7 & \multicolumn{1}{c}{-}  \\
	\multicolumn{1}{l|}{AFA~\cite{chen2020temporal}} & \multicolumn{1}{c|}{ECCV20} & \multicolumn{1}{c|}{Res50} & 82.9 & 90.2 & 96.6 & \multicolumn{1}{c|}{-} & 95.4 & 97.2 & 99.4 & \multicolumn{1}{c|}{99.7} & 88.5 & \multicolumn{1}{c}{96.8}  \\ \midrule
	\multicolumn{1}{l|}{Ours} & \multicolumn{1}{c|}{-} & \multicolumn{1}{c|}{Res50} & \textbf{87.0} & \textbf{90.8} & \textbf{97.1} & \multicolumn{1}{c|}{\textbf{98.8}} & \textbf{97.1} & \textbf{97.6} & \textbf{99.7} & \multicolumn{1}{c|}{\textbf{99.9}}  & \textbf{92.0} & \multicolumn{1}{c}{\textbf{98.0}}   \\ 
	\bottomrule[1.5pt]
	\end{tabular}
	\caption{We compare the DenseIL with state-of-the-art results.}
	\vspace{-3mm}
	\label{tab:sota}
\end{table*}

\subsection{Study on Model Variations}
\label{exp:ablation}

\paragraph{Deeper DI Decoder Improves the Performance.}
As described in Section~\ref{sec:method}, our DI decoder is stacked with $R$ basic building blocks. Correspondingly, we vary different number of blocks to investigate how our model performs. Table~\ref{tab:numofblocks} illustrates the performance (\%) on MARS dataset with $R=0/1/2/3/4$ blocks. We can observe a consistent performance improvement as we increase the number of blocks in DI decoder.

\vspace{-2.5mm}
\paragraph{Wider DI Decoder Improves the Performance.}
We vary dimension of hidden states $d$ to discover how the width of DI decoder affects the re-ID performance. Table~\ref{tab:numofhidden} demonstrates the corresponding results, which tell that wider DI decoder indeed shows larger capability, resulting consistent performance gain on MARS dataset.

\vspace{-2.5mm}
\paragraph{Horizontal Partition Helps Our DI Decoder to Learn Discriminative Features.}
Part-based techniques for person re-ID has remained attractive in the last few years~\cite{sun2018beyond,fu2019sta,yan2020learning,yang2020spatial,zhang2020feature}, we here investigate how horizontal partition influence the learning of DenseIL. The results are illustrated in Table~\ref{tab:numofpart}, from which we can conclude that the best performance is reached when we partition each spatial feature into four patches. Interestingly but not surprisingly, training with larger number of partitions does not achieve higher result (-0.4\% mAP). This might be due to more partitions amplify the misalignment issue existing in video-based person re-ID.

\vspace{-2mm}
\paragraph{Our DI Decoder is not Limited to Specific CNN Backbone.}
For example, DenseIL is able to boost the mAP of DenseNet-121 baseline from 82.5\% to 86.7\% on MARS dataset, demonstrating great generalization ability on different CNN basic structure.

\subsection{Computational Complexity}

For all settings, we compute the theoretical GFLOPS with the tool \texttt{compute\_flops.py} \footnote{\url{https://gist.github.com/fmassa/c0fbb9fe7bf53b533b5cc241f5c8234c}} from DETR~\cite{carion2020end}. From Table~\ref{tab:numofblocks} and~\ref{tab:numofhidden} we can observe that, the deeper and the wider the DI decoder is, the higher computational complexity we have. However, such issue can be alleviated with the recently developed high efficiency Transformer variants~\cite{kitaev2019reformer,zhu2020deformable}. In addition, for the lightweight version in Table~\ref{tab:numofhidden}, \eg, $d=256$, which brings negligible computational overhead, still outperforms the baseline by 4.5\% mAP.

\subsection{Comparison with State-of-the-Art Results}

We compare our proposed DenseIL with the best results reported in recent literatures in Table~\ref{tab:sota}. We can see that our method achieves significant improvement over all the competitive state-of-the-arts, including the optical flow-based~\cite{chen2020temporal}, graph-based~\cite{yang2020spatial,yan2020learning}, 3D CNN-based~\cite{li2019multi,gu2020appearance} and attention-based~\cite{fu2019sta,hou2019vrstc,li2019global,subramaniam2019co,hou2020temporal} approaches, in terms of both the mAP and CMC metrics. For example, our method surpasses the latest optical flow-based scheme~\cite{chen2020temporal} by 4.1\% mAP, outperforms the latest graph-based scheme~\cite{yang2020spatial} by 3.3\% mAP on MARS dataset. For attention-based approaches~\cite{li2019global,hou2020temporal}, our method is also superior to all of them by a large margin. We argue that the performance gain comes from the usage of both the multi-grained cues and the spatial-temporal interaction, which are not fully exploited in previous works.

\section{Conclusion}
In this paper, we successfully leverage Dense Interaction Learning (DenseIL) to alleviate the difficulties of multi-grained spatial-temporal interaction modeling for video-based person re-ID. Specifically, by incorporating the proposed Dense Attention and STEP-Emb, we let the DI decoder densely attends to intermediate CNN features and generates an intrinsically multi-grained representation for each video clip. Experimental results demonstrate that our approach surpasses all previous methods.

For future works, it is interesting to apply Dense Interaction Learning to more video understanding tasks, such as action recognition, video captioning, \etc.

\vspace{-1mm}
\paragraph{Acknowledgements.} This work was supported in part by the National Key Research and Development Program of China 2018AAA0101400 and NSFC Grant U1908209, 61632001, 62021001. We thank all the anonymous reviewers for their valuable comments on our paper.

{\small
\bibliographystyle{ieee_fullname}
\bibliography{egbib}

\begin{thebibliography}{10}\itemsep=-1pt

\bibitem{ba2016layer}
Jimmy~Lei Ba, Jamie~Ryan Kiros, and Geoffrey~E Hinton.
\newblock Layer normalization.
\newblock {\em arXiv preprint arXiv:1607.06450}, 2016.

\bibitem{bahdanau2015neural}
Dzmitry Bahdanau, Kyunghyun Cho, and Yoshua Bengio.
\newblock Neural machine translation by jointly learning to align and
  translate.
\newblock In {\em 3rd International Conference on Learning Representations,
  ICLR 2015}, 2015.

\bibitem{bello2019attention}
Irwan Bello, Barret Zoph, Ashish Vaswani, Jonathon Shlens, and Quoc~V Le.
\newblock Attention augmented convolutional networks.
\newblock In {\em Proceedings of the IEEE International Conference on Computer
  Vision}, pages 3286--3295, 2019.

\bibitem{cao2019gcnet}
Yue Cao, Jiarui Xu, Stephen Lin, Fangyun Wei, and Han Hu.
\newblock Gcnet: Non-local networks meet squeeze-excitation networks and
  beyond.
\newblock In {\em Proceedings of the IEEE International Conference on Computer
  Vision Workshops}, pages 0--0, 2019.

\bibitem{carion2020end}
Nicolas Carion, Francisco Massa, Gabriel Synnaeve, Nicolas Usunier, Alexander
  Kirillov, and Sergey Zagoruyko.
\newblock End-to-end object detection with transformers.
\newblock In {\em European Conference on Computer Vision}, 2020.

\bibitem{chan2016listen}
William Chan, Navdeep Jaitly, Quoc Le, and Oriol Vinyals.
\newblock Listen, attend and spell: A neural network for large vocabulary
  conversational speech recognition.
\newblock In {\em 2016 IEEE International Conference on Acoustics, Speech and
  Signal Processing (ICASSP)}, pages 4960--4964. IEEE, 2016.

\bibitem{chang2018multi}
Xiaobin Chang, Timothy~M Hospedales, and Tao Xiang.
\newblock Multi-level factorisation net for person re-identification.
\newblock In {\em Proceedings of the IEEE Conference on Computer Vision and
  Pattern Recognition}, pages 2109--2118, 2018.

\bibitem{chen2018video}
Dapeng Chen, Hongsheng Li, Tong Xiao, Shuai Yi, and Xiaogang Wang.
\newblock Video person re-identification with competitive snippet-similarity
  aggregation and co-attentive snippet embedding.
\newblock In {\em Proceedings of the IEEE Conference on Computer Vision and
  Pattern Recognition}, pages 1169--1178, 2018.

\bibitem{chen2020temporal}
Guangyi Chen, Yongming Rao, Jiwen Lu, and Jie Zhou.
\newblock Temporal coherence or temporal motion: Which is more critical for
  video-based person re-identification?
\newblock In {\em European Conference on Computer Vision}, 2020.

\bibitem{cho2014learning}
Kyunghyun Cho, Bart van Merri{\"e}nboer, Caglar Gulcehre, Dzmitry Bahdanau,
  Fethi Bougares, Holger Schwenk, and Yoshua Bengio.
\newblock Learning phrase representations using rnn encoder--decoder for
  statistical machine translation.
\newblock In {\em Proceedings of the 2014 Conference on Empirical Methods in
  Natural Language Processing (EMNLP)}, pages 1724--1734, 2014.

\bibitem{chung2017two}
Dahjung Chung, Khalid Tahboub, and Edward~J Delp.
\newblock A two stream siamese convolutional neural network for person
  re-identification.
\newblock In {\em Proceedings of the IEEE International Conference on Computer
  Vision}, pages 1983--1991, 2017.

\bibitem{cornia2020meshed}
Marcella Cornia, Matteo Stefanini, Lorenzo Baraldi, and Rita Cucchiara.
\newblock Meshed-memory transformer for image captioning.
\newblock In {\em Proceedings of the IEEE/CVF Conference on Computer Vision and
  Pattern Recognition}, pages 10578--10587, 2020.

\bibitem{deng2009imagenet}
Jia Deng, Wei Dong, Richard Socher, Li-Jia Li, Kai Li, and Li Fei-Fei.
\newblock Imagenet: A large-scale hierarchical image database.
\newblock In {\em 2009 IEEE conference on computer vision and pattern
  recognition}, pages 248--255. Ieee, 2009.

\bibitem{devlin2019bert}
Jacob Devlin, Ming-Wei Chang, Kenton Lee, and Kristina Toutanova.
\newblock Bert: Pre-training of deep bidirectional transformers for language
  understanding.
\newblock In {\em NAACL-HLT}, 2019.

\bibitem{fu2017dssd}
Cheng-Yang Fu, Wei Liu, Ananth Ranga, Ambrish Tyagi, and Alexander~C Berg.
\newblock Dssd: Deconvolutional single shot detector.
\newblock {\em arXiv preprint arXiv:1701.06659}, 2017.

\bibitem{fu2019dual}
Jun Fu, Jing Liu, Haijie Tian, Yong Li, Yongjun Bao, Zhiwei Fang, and Hanqing
  Lu.
\newblock Dual attention network for scene segmentation.
\newblock In {\em Proceedings of the IEEE Conference on Computer Vision and
  Pattern Recognition}, pages 3146--3154, 2019.

\bibitem{fu2019sta}
Yang Fu, Xiaoyang Wang, Yunchao Wei, and Thomas Huang.
\newblock Sta: Spatial-temporal attention for large-scale video-based person
  re-identification.
\newblock In {\em Proceedings of the AAAI Conference on Artificial
  Intelligence}, volume~33, pages 8287--8294, 2019.

\bibitem{fu2019horizontal}
Yang Fu, Yunchao Wei, Yuqian Zhou, Honghui Shi, Gao Huang, Xinchao Wang,
  Zhiqiang Yao, and Thomas Huang.
\newblock Horizontal pyramid matching for person re-identification.
\newblock In {\em Proceedings of the AAAI Conference on Artificial
  Intelligence}, volume~33, pages 8295--8302, 2019.

\bibitem{gu2018non}
Jiatao Gu, James Bradbury, Caiming Xiong, Victor~OK Li, and Richard Socher.
\newblock Non-autoregressive neural machine translation.
\newblock In {\em International Conference on Learning Representations}, 2018.

\bibitem{gu2020appearance}
Xinqian Gu, Hong Chang, Bingpeng Ma, Hongkai Zhang, and Xilin Chen.
\newblock Appearance-preserving 3d convolution for video-based person
  re-identification.
\newblock In {\em European Conference on Computer Vision}. Springer, 2020.

\bibitem{he2016deep}
Kaiming He, Xiangyu Zhang, Shaoqing Ren, and Jian Sun.
\newblock Deep residual learning for image recognition.
\newblock In {\em Proceedings of the IEEE conference on computer vision and
  pattern recognition}, pages 770--778, 2016.

\bibitem{he2021partial}
Tianyu He, Xu Shen, Jianqiang Huang, Zhibo Chen, and Xian-Sheng Hua.
\newblock Partial person re-identification with part-part correspondence
  learning.
\newblock In {\em Proceedings of the IEEE/CVF Conference on Computer Vision and
  Pattern Recognition}, pages 9105--9115, 2021.

\bibitem{he2018layer}
Tianyu He, Xu Tan, Yingce Xia, Di He, Tao Qin, Zhibo Chen, and Tie-Yan Liu.
\newblock Layer-wise coordination between encoder and decoder for neural
  machine translation.
\newblock In {\em Advances in Neural Information Processing Systems}, pages
  7944--7954, 2018.

\bibitem{hermans2017defense}
Alexander Hermans, Lucas Beyer, and Bastian Leibe.
\newblock In defense of the triplet loss for person re-identification.
\newblock {\em arXiv preprint arXiv:1703.07737}, 2017.

\bibitem{hochreiter1997long}
Sepp Hochreiter and J{\"u}rgen Schmidhuber.
\newblock Long short-term memory.
\newblock {\em Neural computation}, 9(8):1735--1780, 1997.

\bibitem{hou2020temporal}
Ruibing Hou, Hong Chang, Bingpeng Ma, Shiguang Shan, and Xilin Chen.
\newblock Temporal complementary learning for video person re-identification.
\newblock 2020.

\bibitem{hou2019vrstc}
Ruibing Hou, Bingpeng Ma, Hong Chang, Xinqian Gu, Shiguang Shan, and Xilin
  Chen.
\newblock Vrstc: Occlusion-free video person re-identification.
\newblock In {\em Proceedings of the IEEE Conference on Computer Vision and
  Pattern Recognition}, pages 7183--7192, 2019.

\bibitem{hu2019local}
Han Hu, Zheng Zhang, Zhenda Xie, and Stephen Lin.
\newblock Local relation networks for image recognition.
\newblock In {\em Proceedings of the IEEE International Conference on Computer
  Vision}, pages 3464--3473, 2019.

\bibitem{hu2018gather}
Jie Hu, Li Shen, Samuel Albanie, Gang Sun, and Andrea Vedaldi.
\newblock Gather-excite: Exploiting feature context in convolutional neural
  networks.
\newblock In {\em Advances in neural information processing systems}, pages
  9401--9411, 2018.

\bibitem{hu2018squeeze}
Jie Hu, Li Shen, and Gang Sun.
\newblock Squeeze-and-excitation networks.
\newblock In {\em Proceedings of the IEEE conference on computer vision and
  pattern recognition}, pages 7132--7141, 2018.

\bibitem{huang2017densely}
Gao Huang, Zhuang Liu, Laurens Van Der~Maaten, and Kilian~Q Weinberger.
\newblock Densely connected convolutional networks.
\newblock In {\em Proceedings of the IEEE conference on computer vision and
  pattern recognition}, pages 4700--4708, 2017.

\bibitem{jin2020style}
Xin Jin, Cuiling Lan, Wenjun Zeng, Zhibo Chen, and Li Zhang.
\newblock Style normalization and restitution for generalizable person
  re-identification.
\newblock In {\em Proceedings of the IEEE/CVF Conference on Computer Vision and
  Pattern Recognition}, pages 3143--3152, 2020.

\bibitem{jin2020semantics}
Xin Jin, Cuiling Lan, Wenjun Zeng, Guoqiang Wei, and Zhibo Chen.
\newblock Semantics-aligned representation learning for person
  re-identification.
\newblock In {\em Proceedings of the AAAI Conference on Artificial
  Intelligence}, volume~34, pages 11173--11180, 2020.

\bibitem{kipf2016semi}
Thomas~N Kipf and Max Welling.
\newblock Semi-supervised classification with graph convolutional networks.
\newblock In {\em International Conference on Learning Representations}, 2016.

\bibitem{kitaev2019reformer}
Nikita Kitaev, Lukasz Kaiser, and Anselm Levskaya.
\newblock Reformer: The efficient transformer.
\newblock In {\em International Conference on Learning Representations}, 2019.

\bibitem{krizhevsky2012imagenet}
Alex Krizhevsky, Ilya Sutskever, and Geoffrey~E Hinton.
\newblock Imagenet classification with deep convolutional neural networks.
\newblock In {\em Advances in neural information processing systems}, pages
  1097--1105, 2012.

\bibitem{lecun1998gradient}
Yann LeCun, L{\'e}on Bottou, Yoshua Bengio, and Patrick Haffner.
\newblock Gradient-based learning applied to document recognition.
\newblock {\em Proceedings of the IEEE}, 86(11):2278--2324, 1998.

\bibitem{li2017learning}
Dangwei Li, Xiaotang Chen, Zhang Zhang, and Kaiqi Huang.
\newblock Learning deep context-aware features over body and latent parts for
  person re-identification.
\newblock In {\em Proceedings of the IEEE conference on computer vision and
  pattern recognition}, pages 384--393, 2017.

\bibitem{li2019global}
Jianing Li, Jingdong Wang, Qi Tian, Wen Gao, and Shiliang Zhang.
\newblock Global-local temporal representations for video person
  re-identification.
\newblock In {\em Proceedings of the IEEE International Conference on Computer
  Vision}, pages 3958--3967, 2019.

\bibitem{li2019multi}
Jianing Li, Shiliang Zhang, and Tiejun Huang.
\newblock Multi-scale 3d convolution network for video based person
  re-identification.
\newblock In {\em Proceedings of the AAAI Conference on Artificial
  Intelligence}, volume~33, pages 8618--8625, 2019.

\bibitem{li2018diversity}
Shuang Li, Slawomir Bak, Peter Carr, and Xiaogang Wang.
\newblock Diversity regularized spatiotemporal attention for video-based person
  re-identification.
\newblock In {\em Proceedings of the IEEE Conference on Computer Vision and
  Pattern Recognition}, pages 369--378, 2018.

\bibitem{li2014deepreid}
Wei Li, Rui Zhao, Tong Xiao, and Xiaogang Wang.
\newblock Deepreid: Deep filter pairing neural network for person
  re-identification.
\newblock In {\em Proceedings of the IEEE conference on computer vision and
  pattern recognition}, pages 152--159, 2014.

\bibitem{li2015multi}
Xiang Li, Wei-Shi Zheng, Xiaojuan Wang, Tao Xiang, and Shaogang Gong.
\newblock Multi-scale learning for low-resolution person re-identification.
\newblock In {\em Proceedings of the IEEE International Conference on Computer
  Vision}, pages 3765--3773, 2015.

\bibitem{lin2017structured}
Zhouhan Lin, Minwei Feng, Cicero Nogueira~dos Santos, Mo Yu, Bing Xiang, Bowen
  Zhou, and Yoshua Bengio.
\newblock A structured self-attentive sentence embedding.
\newblock In {\em International Conference on Learning Representations}, 2017.

\bibitem{liu2019spatially}
Chih-Ting Liu, Chih-Wei Wu, Yu-Chiang~Frank Wang, and Shao-Yi Chien.
\newblock Spatially and temporally efficient non-local attention network for
  video-based person re-identification.
\newblock In {\em BMVC}, 2019.

\bibitem{liu2017quality}
Yu Liu, Junjie Yan, and Wanli Ouyang.
\newblock Quality aware network for set to set recognition.
\newblock In {\em Proceedings of the IEEE Conference on Computer Vision and
  Pattern Recognition}, pages 5790--5799, 2017.

\bibitem{liu2019spatial}
Yiheng Liu, Zhenxun Yuan, Wengang Zhou, and Houqiang Li.
\newblock Spatial and temporal mutual promotion for video-based person
  re-identification.
\newblock In {\em Proceedings of the AAAI Conference on Artificial
  Intelligence}, volume~33, pages 8786--8793, 2019.

\bibitem{long2015fully}
Jonathan Long, Evan Shelhamer, and Trevor Darrell.
\newblock Fully convolutional networks for semantic segmentation.
\newblock In {\em Proceedings of the IEEE conference on computer vision and
  pattern recognition}, pages 3431--3440, 2015.

\bibitem{lu2019vilbert}
Jiasen Lu, Dhruv Batra, Devi Parikh, and Stefan Lee.
\newblock Vilbert: Pretraining task-agnostic visiolinguistic representations
  for vision-and-language tasks.
\newblock In {\em Advances in Neural Information Processing Systems}, pages
  13--23, 2019.

\bibitem{mao2016image}
Xiaojiao Mao, Chunhua Shen, and Yu-Bin Yang.
\newblock Image restoration using very deep convolutional encoder-decoder
  networks with symmetric skip connections.
\newblock In {\em Advances in neural information processing systems}, pages
  2802--2810, 2016.

\bibitem{mclaughlin2016recurrent}
Niall McLaughlin, Jesus~Martinez Del~Rincon, and Paul Miller.
\newblock Recurrent convolutional network for video-based person
  re-identification.
\newblock In {\em Proceedings of the IEEE conference on computer vision and
  pattern recognition}, pages 1325--1334, 2016.

\bibitem{parikh2016decomposable}
Ankur Parikh, Oscar T{\"a}ckstr{\"o}m, Dipanjan Das, and Jakob Uszkoreit.
\newblock A decomposable attention model for natural language inference.
\newblock In {\em Proceedings of the 2016 Conference on Empirical Methods in
  Natural Language Processing}, pages 2249--2255, 2016.

\bibitem{parmar2019stand}
Niki Parmar, Prajit Ramachandran, Ashish Vaswani, Irwan Bello, Anselm Levskaya,
  and Jon Shlens.
\newblock Stand-alone self-attention in vision models.
\newblock In {\em Advances in Neural Information Processing Systems}, pages
  68--80, 2019.

\bibitem{qian2017multi}
Xuelin Qian, Yanwei Fu, Yu-Gang Jiang, Tao Xiang, and Xiangyang Xue.
\newblock Multi-scale deep learning architectures for person re-identification.
\newblock In {\em Proceedings of the IEEE International Conference on Computer
  Vision}, pages 5399--5408, 2017.

\bibitem{qiu2017learning}
Zhaofan Qiu, Ting Yao, and Tao Mei.
\newblock Learning spatio-temporal representation with pseudo-3d residual
  networks.
\newblock In {\em proceedings of the IEEE International Conference on Computer
  Vision}, pages 5533--5541, 2017.

\bibitem{rahman2020integrating}
Wasifur Rahman, Md~Kamrul Hasan, Sangwu Lee, AmirAli~Bagher Zadeh, Chengfeng
  Mao, Louis-Philippe Morency, and Ehsan Hoque.
\newblock Integrating multimodal information in large pretrained transformers.
\newblock In {\em Proceedings of the 58th Annual Meeting of the Association for
  Computational Linguistics}, pages 2359--2369, 2020.

\bibitem{ristani2016performance}
Ergys Ristani, Francesco Solera, Roger Zou, Rita Cucchiara, and Carlo Tomasi.
\newblock Performance measures and a data set for multi-target, multi-camera
  tracking.
\newblock In {\em European Conference on Computer Vision}, pages 17--35.
  Springer, 2016.

\bibitem{ronneberger2015u}
Olaf Ronneberger, Philipp Fischer, and Thomas Brox.
\newblock U-net: Convolutional networks for biomedical image segmentation.
\newblock In {\em International Conference on Medical image computing and
  computer-assisted intervention}, pages 234--241. Springer, 2015.

\bibitem{su2017pose}
Chi Su, Jianing Li, Shiliang Zhang, Junliang Xing, Wen Gao, and Qi Tian.
\newblock Pose-driven deep convolutional model for person re-identification.
\newblock In {\em Proceedings of the IEEE international conference on computer
  vision}, pages 3960--3969, 2017.

\bibitem{su2019vl}
Weijie Su, Xizhou Zhu, Yue Cao, Bin Li, Lewei Lu, Furu Wei, and Jifeng Dai.
\newblock Vl-bert: Pre-training of generic visual-linguistic representations.
\newblock In {\em International Conference on Learning Representations}, 2019.

\bibitem{subramaniam2019co}
Arulkumar Subramaniam, Athira Nambiar, and Anurag Mittal.
\newblock Co-segmentation inspired attention networks for video-based person
  re-identification.
\newblock In {\em Proceedings of the IEEE International Conference on Computer
  Vision}, pages 562--572, 2019.

\bibitem{sun2019videobert}
Chen Sun, Austin Myers, Carl Vondrick, Kevin Murphy, and Cordelia Schmid.
\newblock Videobert: A joint model for video and language representation
  learning.
\newblock In {\em Proceedings of the IEEE International Conference on Computer
  Vision}, pages 7464--7473, 2019.

\bibitem{sun2018beyond}
Yifan Sun, Liang Zheng, Yi Yang, Qi Tian, and Shengjin Wang.
\newblock Beyond part models: Person retrieval with refined part pooling (and a
  strong convolutional baseline).
\newblock In {\em Proceedings of the European Conference on Computer Vision
  (ECCV)}, pages 480--496, 2018.

\bibitem{tong2017image}
Tong Tong, Gen Li, Xiejie Liu, and Qinquan Gao.
\newblock Image super-resolution using dense skip connections.
\newblock In {\em Proceedings of the IEEE International Conference on Computer
  Vision}, pages 4799--4807, 2017.

\bibitem{varior2016siamese}
Rahul~Rama Varior, Bing Shuai, Jiwen Lu, Dong Xu, and Gang Wang.
\newblock A siamese long short-term memory architecture for human
  re-identification.
\newblock In {\em European conference on computer vision}, pages 135--153.
  Springer, 2016.

\bibitem{vaswani2017attention}
Ashish Vaswani, Noam Shazeer, Niki Parmar, Jakob Uszkoreit, Llion Jones,
  Aidan~N Gomez, {\L}ukasz Kaiser, and Illia Polosukhin.
\newblock Attention is all you need.
\newblock In {\em Advances in neural information processing systems}, pages
  5998--6008, 2017.

\bibitem{wang2018learning}
Guanshuo Wang, Yufeng Yuan, Xiong Chen, Jiwei Li, and Xi Zhou.
\newblock Learning discriminative features with multiple granularities for
  person re-identification.
\newblock In {\em Proceedings of the 26th ACM international conference on
  Multimedia}, pages 274--282, 2018.

\bibitem{wang2020axial}
Huiyu Wang, Yukun Zhu, Bradley Green, Hartwig Adam, Alan Yuille, and
  Liang-Chieh Chen.
\newblock Axial-deeplab: Stand-alone axial-attention for panoptic segmentation.
\newblock In {\em Proceedings of the European Conference on Computer Vision},
  2020.

\bibitem{wang2014person}
Taiqing Wang, Shaogang Gong, Xiatian Zhu, and Shengjin Wang.
\newblock Person re-identification by video ranking.
\newblock In {\em European conference on computer vision}, pages 688--703.
  Springer, 2014.

\bibitem{wang2013intelligent}
Xiaogang Wang.
\newblock Intelligent multi-camera video surveillance: A review.
\newblock {\em Pattern recognition letters}, 34(1):3--19, 2013.

\bibitem{wang2018non}
Xiaolong Wang, Ross Girshick, Abhinav Gupta, and Kaiming He.
\newblock Non-local neural networks.
\newblock In {\em Proceedings of the IEEE conference on computer vision and
  pattern recognition}, pages 7794--7803, 2018.

\bibitem{wang2019multi}
Yiren Wang, Yingce Xia, Tianyu He, Fei Tian, Tao Qin, ChengXiang Zhai, and
  Tie-Yan Liu.
\newblock Multi-agent dual learning.
\newblock In {\em Proceedings of the International Conference on Learning
  Representations (ICLR) 2019}, 2019.

\bibitem{wei2017glad}
Longhui Wei, Shiliang Zhang, Hantao Yao, Wen Gao, and Qi Tian.
\newblock Glad: Global-local-alignment descriptor for pedestrian retrieval.
\newblock In {\em Proceedings of the 25th ACM international conference on
  Multimedia}, pages 420--428, 2017.

\bibitem{wu2018exploit}
Yu Wu, Yutian Lin, Xuanyi Dong, Yan Yan, Wanli Ouyang, and Yi Yang.
\newblock Exploit the unknown gradually: One-shot video-based person
  re-identification by stepwise learning.
\newblock In {\em Proceedings of the IEEE Conference on Computer Vision and
  Pattern Recognition}, pages 5177--5186, 2018.

\bibitem{xia2019tied}
Yingce Xia, Tianyu He, Xu Tan, Fei Tian, Di He, and Tao Qin.
\newblock Tied transformers: Neural machine translation with shared encoder and
  decoder.
\newblock In {\em Proceedings of the AAAI Conference on Artificial
  Intelligence}, volume~33, pages 5466--5473, 2019.

\bibitem{xiong2020layer}
Ruibin Xiong, Yunchang Yang, Di He, Kai Zheng, Shuxin Zheng, Chen Xing,
  Huishuai Zhang, Yanyan Lan, Liwei Wang, and Tie-Yan Liu.
\newblock On layer normalization in the transformer architecture.
\newblock In {\em Proceedings of the International Conference on Machine
  Learning}, 2020.

\bibitem{xu2017jointly}
Shuangjie Xu, Yu Cheng, Kang Gu, Yang Yang, Shiyu Chang, and Pan Zhou.
\newblock Jointly attentive spatial-temporal pooling networks for video-based
  person re-identification.
\newblock In {\em Proceedings of the IEEE international conference on computer
  vision}, pages 4733--4742, 2017.

\bibitem{yan2016person}
Yichao Yan, Bingbing Ni, Zhichao Song, Chao Ma, Yan Yan, and Xiaokang Yang.
\newblock Person re-identification via recurrent feature aggregation.
\newblock In {\em European Conference on Computer Vision}, pages 701--716.
  Springer, 2016.

\bibitem{yan2020learning}
Yichao Yan, Jie Qin, Jiaxin Chen, Li Liu, Fan Zhu, Ying Tai, and Ling Shao.
\newblock Learning multi-granular hypergraphs for video-based person
  re-identification.
\newblock In {\em Proceedings of the IEEE/CVF Conference on Computer Vision and
  Pattern Recognition}, pages 2899--2908, 2020.

\bibitem{yang2020spatial}
Jinrui Yang, Wei-Shi Zheng, Qize Yang, Ying-Cong Chen, and Qi Tian.
\newblock Spatial-temporal graph convolutional network for video-based person
  re-identification.
\newblock In {\em Proceedings of the IEEE/CVF Conference on Computer Vision and
  Pattern Recognition}, pages 3289--3299, 2020.

\bibitem{yang2019xlnet}
Zhilin Yang, Zihang Dai, Yiming Yang, Jaime Carbonell, Russ~R Salakhutdinov,
  and Quoc~V Le.
\newblock Xlnet: Generalized autoregressive pretraining for language
  understanding.
\newblock In {\em Advances in neural information processing systems}, pages
  5753--5763, 2019.

\bibitem{zeng2020learning}
Yanhong Zeng, Jianlong Fu, and Hongyang Chao.
\newblock Learning joint spatial-temporal transformations for video inpainting.
\newblock In {\em European Conference on Computer Vision}, pages 528--543.
  Springer, 2020.

\bibitem{zhang2020feature}
Dong Zhang, Hanwang Zhang, Jinhui Tang, Meng Wang, Xiansheng Hua, and Qianru
  Sun.
\newblock Feature pyramid transformer.
\newblock In {\em Proceedings of the European Conference on Computer Vision},
  2020.

\bibitem{zhang2019self}
Han Zhang, Ian Goodfellow, Dimitris Metaxas, and Augustus Odena.
\newblock Self-attention generative adversarial networks.
\newblock In {\em International Conference on Machine Learning}, pages
  7354--7363. PMLR, 2019.

\bibitem{zhang2020multi}
Zhizheng Zhang, Cuiling Lan, Wenjun Zeng, and Zhibo Chen.
\newblock Multi-granularity reference-aided attentive feature aggregation for
  video-based person re-identification.
\newblock In {\em Proceedings of the IEEE/CVF Conference on Computer Vision and
  Pattern Recognition}, pages 10407--10416, 2020.

\bibitem{zhao2019attribute}
Yiru Zhao, Xu Shen, Zhongming Jin, Hongtao Lu, and Xian-sheng Hua.
\newblock Attribute-driven feature disentangling and temporal aggregation for
  video person re-identification.
\newblock In {\em Proceedings of the IEEE conference on computer vision and
  pattern recognition}, pages 4913--4922, 2019.

\bibitem{zheng2016mars}
Liang Zheng, Zhi Bie, Yifan Sun, Jingdong Wang, Chi Su, Shengjin Wang, and Qi
  Tian.
\newblock Mars: A video benchmark for large-scale person re-identification.
\newblock In {\em European Conference on Computer Vision}, pages 868--884.
  Springer, 2016.

\bibitem{zheng2015scalable}
Liang Zheng, Liyue Shen, Lu Tian, Shengjin Wang, Jingdong Wang, and Qi Tian.
\newblock Scalable person re-identification: A benchmark.
\newblock In {\em Proceedings of the IEEE international conference on computer
  vision}, pages 1116--1124, 2015.

\bibitem{zheng2017sift}
Liang Zheng, Yi Yang, and Qi Tian.
\newblock Sift meets cnn: A decade survey of instance retrieval.
\newblock {\em IEEE transactions on pattern analysis and machine intelligence},
  40(5):1224--1244, 2017.

\bibitem{zhong2020random}
Zhun Zhong, Liang Zheng, Guoliang Kang, Shaozi Li, and Yi Yang.
\newblock Random erasing data augmentation.
\newblock In {\em AAAI}, pages 13001--13008, 2020.

\bibitem{zhou2019omni}
Kaiyang Zhou, Yongxin Yang, Andrea Cavallaro, and Tao Xiang.
\newblock Omni-scale feature learning for person re-identification.
\newblock In {\em Proceedings of the IEEE International Conference on Computer
  Vision}, pages 3702--3712, 2019.

\bibitem{zhou2017see}
Zhen Zhou, Yan Huang, Wei Wang, Liang Wang, and Tieniu Tan.
\newblock See the forest for the trees: Joint spatial and temporal recurrent
  neural networks for video-based person re-identification.
\newblock In {\em Proceedings of the IEEE Conference on Computer Vision and
  Pattern Recognition}, pages 4747--4756, 2017.

\bibitem{zhu2017unpaired}
Jun-Yan Zhu, Taesung Park, Phillip Isola, and Alexei~A Efros.
\newblock Unpaired image-to-image translation using cycle-consistent
  adversarial networks.
\newblock In {\em Proceedings of the IEEE international conference on computer
  vision}, pages 2223--2232, 2017.

\bibitem{zhu2020deformable}
Xizhou Zhu, Weijie Su, Lewei Lu, Bin Li, Xiaogang Wang, and Jifeng Dai.
\newblock Deformable detr: Deformable transformers for end-to-end object
  detection.
\newblock {\em arXiv preprint arXiv:2010.04159}, 2020.

\end{thebibliography}
}

\end{document}